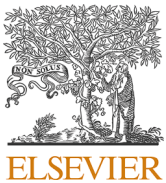
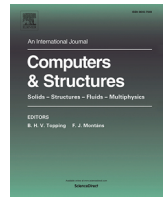

# Transformer self-attention encoder–decoder with multimodal deep learning for response time series forecasting and digital twin support in wind structural health monitoring

Feiyu Zhou [a,b], Marios Impraimakis [a,*]

[a] *Department of Mechanical Engineering, University of Bath, BA2 7AY Bath, UK*
[b] *Department of Civil Engineering, Zhejiang University, Hangzhou 310058, China*

## HIGHLIGHTS

- A transformer model is introduced for wind-excited structural response forecasting.
- The multimodal network jointly learns from wind features and vibration signals.
- Real-world Hardanger Bridge data validate the model under variable conditions.
- Contributes to modal energy retention, tail-risk reduction, and fewer false alarms.

ABSTRACT

The wind-induced structural response forecasting capabilities of a novel transformer methodology are examined here. The model also provides a digital twin component for bridge structural health monitoring. Firstly, the approach uses the temporal characteristics of the system to train a forecasting model. Secondly, the vibration predictions are compared to the measured ones to detect large deviations. Finally, the identified cases are used as an early-warning indicator of structural change. The artificial intelligence-based model outperforms approaches for response forecasting as no assumption on wind stationarity or on structural normal vibration behavior is needed. Specifically, wind-excited dynamic behavior suffers from uncertainty related to obtaining poor predictions when the environmental or traffic conditions change. This results in a hard distinction of what constitutes normal vibration behavior. To this end, a framework is rigorously examined on real-world measurements from the Hardanger Bridge monitored by the Norwegian University of Science and Technology. The approach captures accurate structural behavior in realistic conditions, and with respect to the changes in the system excitation. The results, importantly, highlight the potential of transformer-based digital twin components to serve as next-generation tools for resilient infrastructure management, continuous learning, and adaptive monitoring over the system's lifecycle with respect to temporal characteristics.

## 1. Introduction

Response forecasting and digital twinning play a central role in dynamics-based structural health monitoring. In particular, the behavior of structural systems described by Wagg et al. [1,2] is not only uncertain but also changes over time, often under diverse environmental influences [3–6]. Incorporating data–driven modeling into digital twin frameworks strengthens real-time monitoring, predictive maintenance, and structural health management [7,8], among various engineering applications [9,10]. While output-only identification approaches within digital twin systems serve as powerful tools for realistic civil applications [11–15], joint input–output methods tend to yield more accurate results. These approaches lead to well-defined digital twins and decision-making processes in structural health monitoring [16,17].

On the deep learning infrastructure engineering front [18–24], approaches have addressed complex problems such as crack segmentation [25,26], and underwater inspection tasks [27]. At the signal level, deep sparse autoencoders and sparse Bayesian learning methods have been employed to extract damage-sensitive representations with greater interpretability by Pathirage et al. [28], while meta-learning has been explored to promote cross-task generalization by Xu et al. [29]. Moreover, L1-based sparse regularization has been utilized to improve automatic damage localization by Bao et al. [30]. Collectively, these advancements illustrate that deep learning can effectively integrate multi-sensor information, manage nonlinear behaviors, and enhance automation in processes such as data cleaning, modal identification, and damage localization [31,32].

Transformers, originally developed for sequence modeling through self-attention by Vaswani et al. [33], have recently gained interest in structural health monitoring due to their capacity to capture long-term dependencies and enable sequence-to-sequence prediction [34–38]. Vision-oriented transformer architectures have also been adapted for high-resolution imagery in structural health monitoring, facilitating detailed segmentation and feature extraction, as seen by Azimi et al. [36]. For structural time-series analysis, hybrid recurrent–transformer frameworks have been introduced to improve deformation prediction through temporal correction mechanisms proposed by Wang et al. [39], and transformer-based denoising has been shown to enhance mechanical vibration signals for subsequent anomaly detection, as examined by Chen et al. [40]. Unsupervised and weakly supervised variants, including temporal fusion transformers and related attention-based fusion methods, have also been explored by Falchi et al. [37] for anomaly detection in heritage structures. Beyond task-specific models, recent efforts have established foundation-model paradigms in structural health monitoring [41–46], leveraging masked-autoencoder transformers pretrained on extensive bridge datasets and fine-tuned for applications such as anomaly detection and traffic/load estimation [43].

Transformer networks, specifically, employ self-attention mechanisms to capture long-range dependencies and cross-modal relationships without relying on manually crafted features [47–52]. In contrast, traditional structural health monitoring approaches depend largely on linear, hand-engineered signal-processing routines, making them less suitable for nonlinear or long-range dynamic behaviors [53]. As systems scale to heterogeneous, noisy, and non-stationary data, classical methods often require significant manual feature tuning and threshold selection, leading to false alarms or undetected minor damages [32,53,54]. To address these limitations, data-driven deep learning frameworks have gained popularity by directly learning complex, damage-sensitive representations from raw or minimally processed signals, as described by Bao and Li [31]. For preprocessing and data cleaning, stacked autoencoders and convolutional neural network-based classifiers have been adopted to identify abnormal patterns and classify time—frequency maps of structural responses, thereby improving the reliability of outlier removal and data recovery. Transfer learning techniques have been leveraged to transfer knowledge between datasets and structures, thus enhancing generalization even with limited labeled data, with an application by Pan et al. [55]. Class imbalance issues can be alleviated through tailored loss-function designs [56,57]. For modal parameter estimation in particular, recent studies have complemented or replaced classical methods with machine learning algorithms to improve automation and robustness to noise [58,59]. Furthermore, recent developments include mechanics-informed neural networks that exploit time—frequency sparsity and modal independence to achieve fully automated parameter identification [60].

The accurate representation of wind-induced dynamics under turbulent conditions continues to pose major challenges in wind engineering, though. Recent progress in artificial intelligence and machine learning, with or without explicit physical modeling, has introduced new pathways for simulating engineering phenomena, particularly within wind-related structural health monitoring [61–63]. In a broader context, Arul et al. [64] proposed a data-driven technique that autonomously identifies thunderstorms from large datasets of high-frequency wind speed records, independent of wind statistical parameters. Lin et al. [65] developed a method for predicting crosswind force spectra and the corresponding responses of tall rectangular buildings, with several other contributions expanding this field [66–69]. The research remains active across both local and global scales [70–74]. Within the domain of generative artificial intelligence [75] for wind engineering, Ye et al. [76] introduced an auxiliary classification time-series generative adversarial network to emulate day-ahead wind power fluctuations and output levels, while Liu et al. [77] proposed a graph-based generative adversarial model for predicting short-term wind field evolution, reflecting a growing research interest [78–81]. In the context of wind—structure interaction, Rozov and Breitsamter [82] constructed a reduced-order model to forecast motion-induced unsteady pressure distributions, Mentzelopoulos et al. [83] employed a transformer-based neural network to predict vortex-induced vibrations in flexible cylinders, and Lim et al. [84] compared resistive force theory against advanced slender-body approaches, proposing a neural hydrodynamic model that leverages local drag coefficients for efficient computation.

However, existing wind-induced dynamics learning models often fail to capture the long-term temporal dependencies inherent in structural dynamics [85]. This study introduces a new transformer neural network framework designed to address these challenges. The proposed model simultaneously learns the temporal characteristics of the structure and its interactions with wind–excitation. It outperforms existing forecasting approaches since it does not require assumptions regarding wind stationarity or normal structural behavior, making it suitable for real-world operational conditions [32,33,54]. The framework is rigorously validated using real-world measurements from the Hardanger Bridge monitored by the Norwegian University of Science and Technology. The results demonstrate that the proposed approach offers a simple, yet powerful and effective, tool for monitoring systems and vibration-based damage detection in wind–structure dynamics.

The work is organized as follows: Section 2 outlines the standard wind–structure interaction modeling framework and the corresponding numerical simulations. Section 3 presents the transformer neural network architecture and its extension to early-warning structural health monitoring and forecasting. Section 4 details the performance metrics used for model evaluation. Section 5 describes the structural system under investigation, followed by the results in Section 6. Section 7 provides discussion, algorithmic details, and directions for future research, while Section 8 concludes the work.

## 2. Wind-bridge interaction modeling

The aerodynamic behavior of the structure illustrated in Fig. 1 is analyzed using a three-dimensional Theodorsen formulation that accounts for the combined effects of aerodynamic forces, structural deformation, and wind–structure interactions [86–88]. The model includes a vertical displacement $h(t)$ representing the bridge deck's motion under aerodynamic excitation, a pitching angle $\phi(t)$ describing the rotation about the elastic axis, and a torsional deformation $p(t)$ denoting the twisting caused by asymmetric aerodynamic loads, expressed as:

$$m\,\ddot{h}(t) + S\,\ddot{\phi}(t) + c_h\,\dot{h}(t) + k_h\,h(t)$$
$$= \frac{1}{2}\rho u^2(t)\,B^2 \left( \Omega H_1 \frac{\dot{h}(t)}{u(t)} + \Omega H_2 \frac{B\dot{\phi}(t)}{u(t)} + \Omega^2 H_3\,\phi(t) + \Omega^2 H_4 \frac{h(t)}{B} \right.$$
$$\left. + \Omega H_5 \frac{\dot{p}(t)}{u(t)} + \Omega^2 H_6 \frac{p(t)}{B} \right) \quad (1)$$

$$I\,\ddot{\phi}(t) + S\,\ddot{h}(t) + c_\phi\,\dot{\phi}(t) + k_\phi\,\phi(t)$$
$$= \frac{1}{2}\rho u^2(t) B^2 \left( \Omega A_1 \frac{\dot{h}(t)}{u(t)} + \Omega A_2 \frac{B\,\dot{\phi}(t)}{u(t)} + \Omega^2 A_3\,\phi(t) + \Omega^2\,A_4 \frac{h(t)}{B} \right.$$
$$\left. + \Omega\,A_5 \frac{\dot{p}(t)}{u(t)} + \Omega^2\,A_6 \frac{p(t)}{B} \right) \quad (2)$$

$$m\,\ddot{p}(t) + c_p\,\dot{p}(t) + k_p\,p(t) = \frac{1}{2}\,\rho\,u^2(t)\,B^2 \left( \Omega P_1 \frac{\dot{p}(t)}{u(t)} + \Omega P_2 \frac{B\dot{\phi}(t)}{u(t)} + \Omega^2 P_3\,\phi(t) \right.$$
$$\left. + \Omega^2 P_4 \frac{p(t)}{B} + \Omega P_5 \frac{\dot{h}(t)}{u(t)} + \Omega^2 P_6 \frac{h(t)}{B} \right) \quad (3)$$

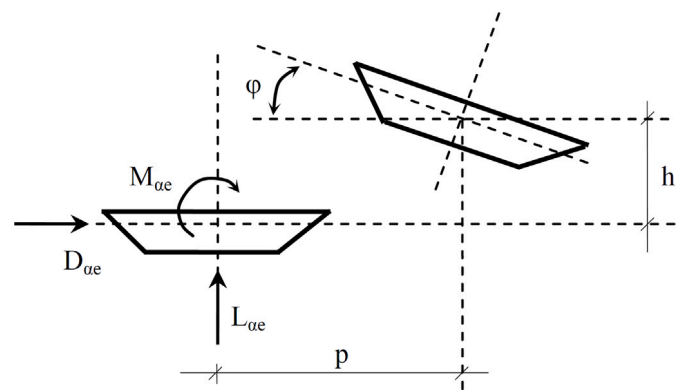

**Fig. 1.** Three-dimensional Theodorsen bridge deck representation of Section 2, incorporating aerodynamic loads, structural deformation, and airflow interactions.

**Table 1**
Parameter values used in the Theodorsen bridge deck model of the Section 2 analysis.

| Parameter | Value |
| --- | --- |
| Mass per unit length, $m$ | 500 kg/m |
| Static moment, $S$ | 100 kg m |
| Damping coefficient, $c_h$ | 250 Ns/m |
| Stiffness coefficient, $k_h$ | $10^4$ N/m |
| Moment of inertia, $I$ | 100 kg m$^2$ |
| Torsional damping coefficient, $c_\phi$ | 250 Ns/rad |
| Torsional stiffness coefficient, $k_\phi$ | $2 \times 10^5$ N m/rad |
| Air density, $\rho$ | 1.29 kg/m$^3$ |
| Characteristic width, $B$ | 12 m |
| Normalized frequency, $W$ | 2.10 |
| Aerodynamic coefficients: | |
| $H_1$ | 1.61 |
| $H_2$ | 1.26 |
| $H_3$ | 1.86 |
| $H_4$ | $-0.957$ |
| $A_1$ | 0.807 |
| $A_2$ | 0.864 |
| $A_3$ | 1.323 |
| $A_4$ | 1.092 |

**Table 2**
Parameters for the turbulent wind velocity field simulation using the spectral representation method [89].

| Parameter | Description | Value |
| --- | --- | --- |
| $z_1$ | Reference height 1 | 20 m |
| $z_0$ | Surface roughness length | 0.001266 |
| $u_s$ | Scale velocity | 1.76 m/s |
| $\omega_u$ | Angular frequency | 4.0 rad/s |
| $N$ | Number of time steps | 2048 |
| $M$ | Number of spatial points | 4096 |

Here, $m$ denotes the mass per unit length, $I$ is the moment of inertia, $c_h$ and $c_\phi$ represent the damping coefficients, $k_h$ and $k_\phi$ are the stiffness constants, $S$ is the static moment, $\rho$ is the air density, $u(t)$ corresponds to the wind velocity, and $B$ is the characteristic deck width. The coefficients $H_{1:6}$, $P_{1:6}$, and $A_{1:6}$ are the aerodynamic flutter derivatives, while $\Omega$ is the normalized oscillation frequency defined with respect to $B$ and the mean wind velocity.

Notably, the first two equations for $h(t)$ and $\phi(t)$ are strongly coupled through the static moment $S$, meaning the vertical heave motion directly influences pitch and vice versa. In contrast, the torsional equation for $p(t)$ excludes $I$ and $S$, suggesting weaker direct coupling, though aerodynamic terms maintain an indirect interdependence with $h(t)$ and $\phi(t)$. To estimate the bridge deck's dynamic behavior, the Runge–Kutta fourth-order integration method is applied over a 12-second duration. The sampling frequency for the measured dynamic states is set at 100 Hz, resulting in a time step $\Delta t = 0.01$ s in the numerical integration. Aerodynamic forces are calculated under a harmonic response assumption to simplify computation, with the $p(t)$ motion neglected without loss of generality. The simulation employs the parameter values summarized in Table 1. Table 2 provides the parameters for the turbulent wind velocity field simulation using the spectral representation method [89], while Table 3 summarizes the stability characteristics of the simulation.

The simulation produces two sets of dynamic responses: one derived from accurate modeling and another one from slightly inaccurate modeling, depending on whether the wind turbulence inputs are precise or imperfect, as seen in Table 3. Fig. 2 presents the estimated vertical displacements, velocities, and rotational responses of the structure, emphasizing how the model estimation errors affect the dynamic response. It is observed that even small inaccuracies in the input data can lead to substantial deviations in dynamic predictions, as seen also in Fig. 3 when comparing the prediction error. The proposed transformer-based model is then examined as a means to address and compensate for such uncertainties.

## 3. A multimodal transformer encoder–decoder for response forecasting and early-warning structural health monitoring

To mathematically describe the forecasting model at a given time $t$, let $\mathbf{w}_t \in \mathbf{R}^{d_w}$ denote the vector of wind features, namely horizontal wind speed, wind direction, turbulence intensity, or temperature. Additionally, let $\mathbf{a}_t \in \mathbb{R}^{d_a}$ denote the vector of structural accelerations at that time. The scope is to provide forecasting for a sliding time window of length $L$ into the future. To this end, the historical window for wind, written as $X_{\text{wind}} = \{\mathbf{w}_{t-L_{\text{enc}}+1}, \ldots, \mathbf{w}_t\} \in \mathbb{R}^{L_{\text{enc}} \times d_w}$, and the historical acceleration window, written as $X_{\text{acc}} = \{\mathbf{a}_{t-L_{\text{enc}}+1}, \ldots, \mathbf{a}_t\} \in \mathbb{R}^{L_{\text{enc}} \times d_a}$, are used as the model input. The final learning task is to predict the future accelerations $\hat{Y} = \{\hat{\mathbf{a}}_{t+1}, \ldots, \hat{\mathbf{a}}_{t+L_{\text{pred}}}\} \in \mathbb{R}^{L_{\text{pred}} \times d_a}$ based on these past windows ($X_{\text{wind}}, X_{\text{acc}}$).

**Table 3**
Stability properties, numerical error estimates, and imperfect wind–input modeling details.

| Quantity | Value | Explanation |
| --- | --- | --- |
| Continuous-time eigenvalues of $A$ | $[-1.6260 \pm 50.0234i,\ -0.2490 \pm 4.4607i]$ | All real parts are negative $\to$ asymptotically stable. |
| Discrete-time eigenvalues of $A_d$ | $[0.9965 \pm 0.0445i,\ 0.8588 \pm 0.4921i]$ | Magnitudes $< 1$, ensuring stable discrete-time propagation. |
| Spectral radius $\rho(A_d)$ | 0.99751 | Confirms $A_d$ is contractive and stable for $\Delta t = 0.01$. |
| Truncation error norm $\|E_{\text{trunc}}\|$ | 1.0603 | Third-order Taylor remainder does not accumulate due to stability. |
| Round-off error magnitude | $2.2204 \times 10^{-14}$ | Double-precision floating point negligible vs. loads (10–100 N). |
| Imperfect wind input | $\tilde{u}(t) = u(t) + 0.2\,\mathcal{N}(0,1) + 3$ | Gaussian noise ($\sigma = 0.2$) and bias ($+3$) models turbulent-input uncertainty. |

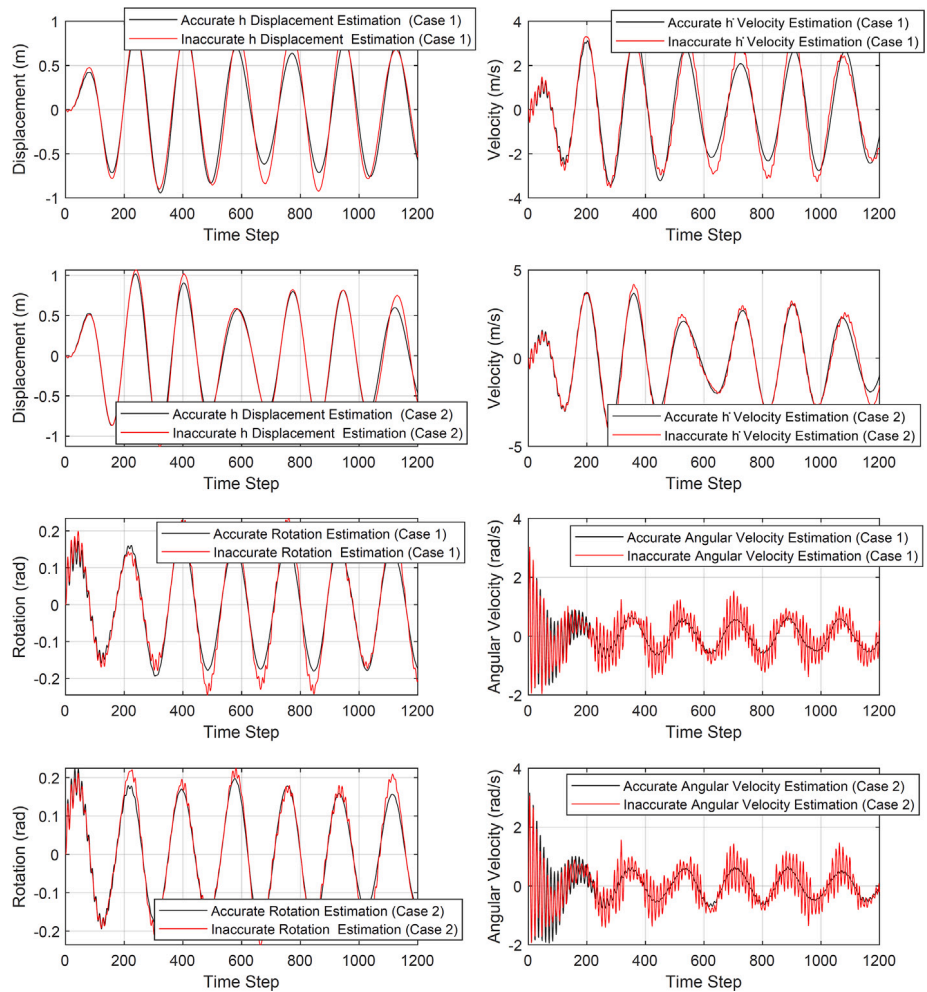

**Fig. 2.** Comparison of aerodynamic response estimations under accurate and perturbed wind conditions in the Section 2 Theodorsen bridge deck model, where each case refers to a different simulation comparison.

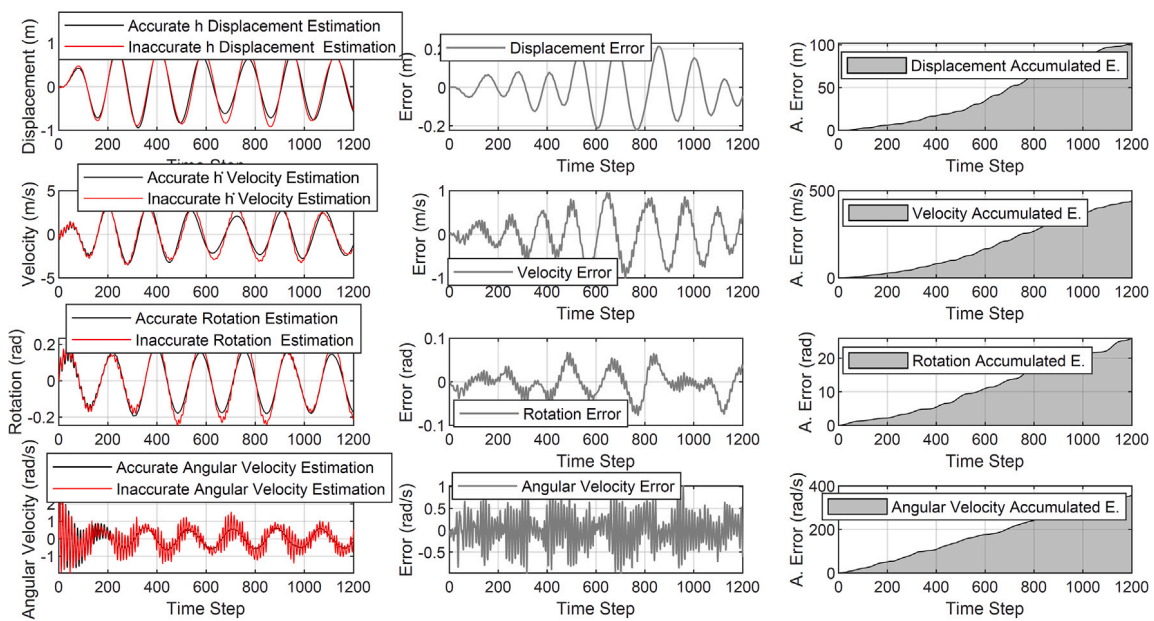

**Fig. 3.** Error and accumulated error for the aerodynamic response estimations under accurate and perturbed wind conditions in the Section 2 Theodorsen bridge deck model.

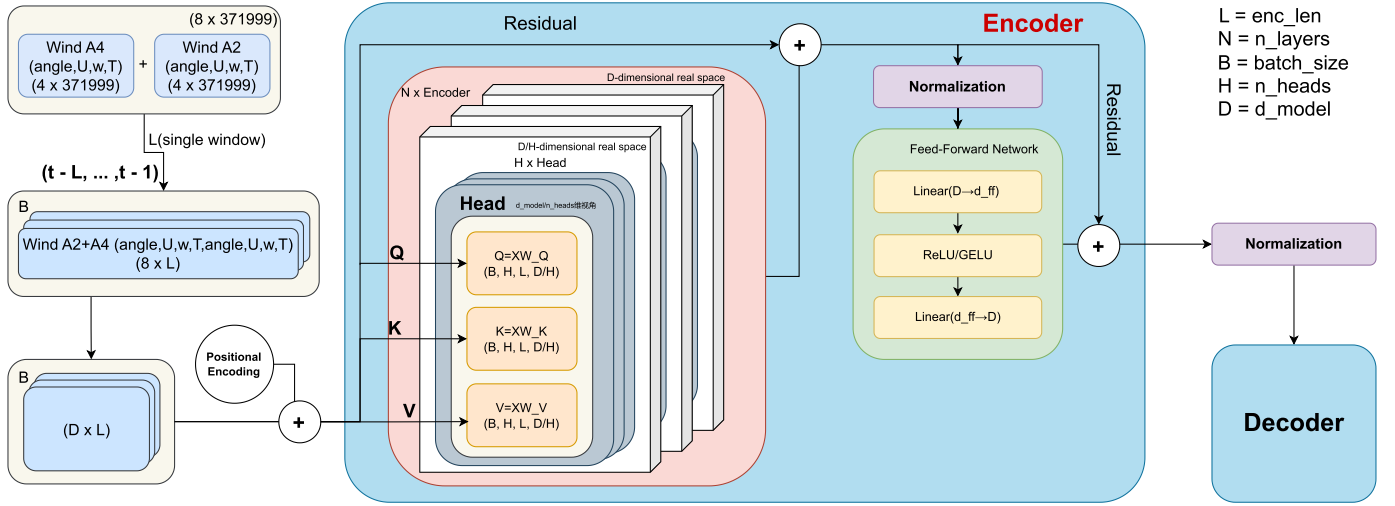

**Fig. 4.** Encoder of the multimodal transformer of Section 3. Wind features are embedded with positional encoding, and passed through N stacked encoder blocks to produce a memory tensor M for cross-modal conditioning in the decoder.

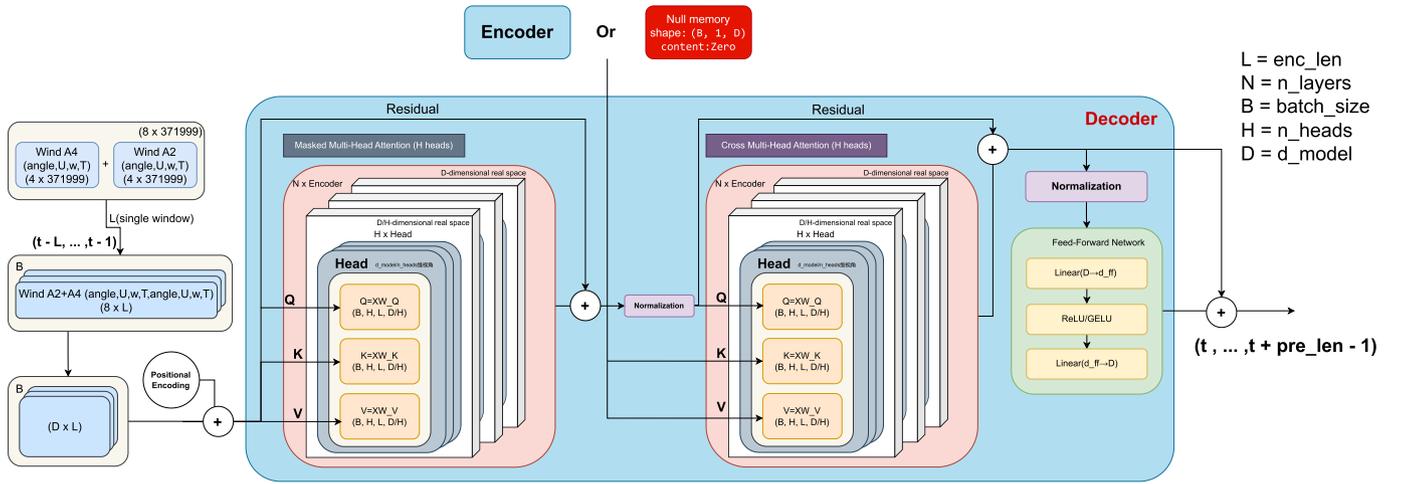

**Fig. 5.** Decoder of the multimodal transformer of Section 3. Given the past acceleration sequence, the decoder applies masked self-attention and cross-attention over the encoder memory M. A linear head maps the decoded states to accelerations, generating autoregressively the future sequence.

For training and validation purposes, the ground-truth target, written as $Y$, consists of the actual future accelerations $\{\mathbf{a}_{t+1}, \ldots, \mathbf{a}_{t+L_{\text{pred}}}\}$. In this work, two input scenarios are investigated: an acceleration-only forecaster (encoder bypassed), and a multimodal forecaster that uses both the wind and the past accelerations. The goal is to forecast the future structural acceleration when training the model with only the history of accelerations, or when training the model with both the wind and the acceleration history recordings. An additional fixed time delay process is also examined by shifting the wind sequence relative to acceleration, to account for known physical lags. This is implemented by zero-padding and shifting the wind array when a positive delay is specified.

The examined encoder–decoder transformer architecture, which jointly processes wind and acceleration histories to forecast future accelerations, has two main components: an encoder sub-network that processes the wind modality, and a decoder sub-network that generates acceleration predictions incorporating both self-attention and cross-attention, as shown in Figs. 4 and 5.

Specifically, the encoder collects wind features into a memory $\mathbf{M}$, and the decoder autoregressively provides the forecasting $\hat{\mathbf{A}}_{(t+1):(t+H)}$ when the history $\mathbf{A}_{\text{hist}}$ is provided. In the acceleration-only training scenario, the encoder is bypassed, and a zero tensor serves as a dummy memory. Each of the $N$ decoder layers comprises a masked multi-head self-attention mechanism over the decoder inputs to enforce causality, a multi-head encoder–decoder attention mechanism over $\mathbf{M}$, and a position-wise feed-forward network. A final linear head maps the top-layer decoder states to $\mathbb{R}^{d_a}$. Importantly, for multi-step forecasting over a larger period of $L_{\text{pred}}$, an iterative autoregressive decoding process is followed. Specifically, starting with $\mathbf{A}_{\text{hist}}$, the decoder outputs $\hat{\mathbf{a}}_{t+1}$ by embedding previous predictions, and the process repeats to produce $\{\hat{\mathbf{a}}_{t+h}\}_{h=1}^{L_{\text{pred}}}$.

The cross-modal mechanism is realized via the encoder–decoder attention, allowing each forecast step to access the entire wind memory $\mathbf{M}$. Specifically, wind and acceleration modalities are integrated through the encoder–decoder cross-attention mechanism in each decoder layer. The decoder's multi-head attention sublayer takes the input from the decoder's previous output, as well as values from the encoder's output memory $\mathbf{M}$. Namely, at each decoder time step, the network "attends to" the entire wind sequence via the encoder memory. This encoder-decoder attention allows each output position to consider all input positions. Consequently, the decoder selectively incorporates wind information when predicting future acceleration. The attention weights (of shape $L_{\text{query}} \times L_{\text{enc}}$) can be interpreted as a cross-modal relevance map between acceleration inputs and wind features. If a strong gust occurred at some past time, the model attends more strongly to that instant to forecast the structural response, capturing the causal influence of wind on vibrations. The multi-head mechanism, therefore, provides several attention heads to do this in parallel, allowing the model to learn multiple alignment patterns.

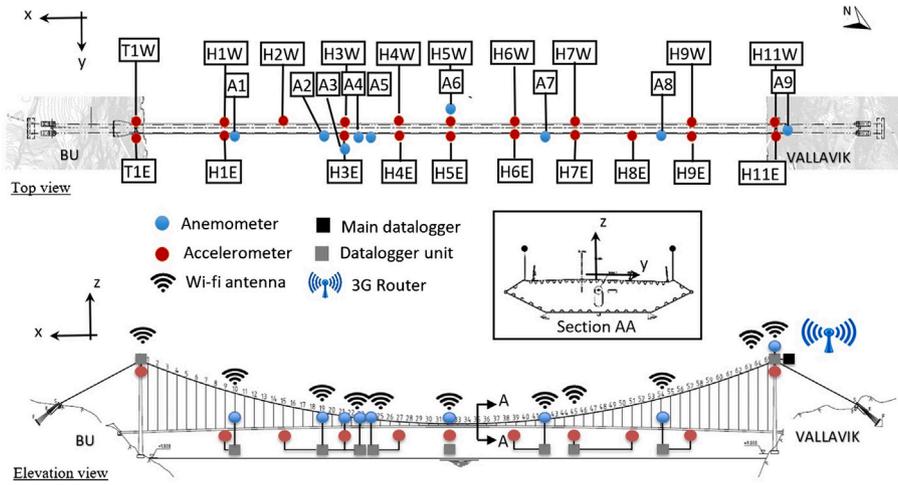

**Fig. 6.** Overview of the Hardanger bridge monitoring system showing the accelerometers (H1–H11), the anemometers (A1–A9), the Wi-Fi/GPS units, and the dataloggers [90].

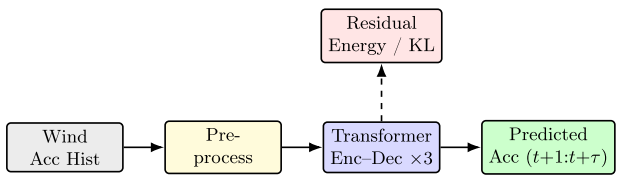

**Fig. 7.** The forecasting stages where wind/acceleration measurements are embedded, processed by stacked transformer blocks, and decoded to predict future bridge accelerations. A final residual-energy module may compute anomaly scores for damage detection.

Specifically, each attention block computes a weighted context representation using the scaled dot–product operation. Given the query $Q$, the key $K$, and the value $V$ matrices of dimension $d_k$, the multi-head attention is written as:

$$\text{Attention}(Q, K, V) = \text{softmax}\left(\frac{QK^\top}{\sqrt{d_k}}\right) V \quad (4)$$

In the multimodal encoder–decoder, the encoder self-attention learns temporal dependencies among the wind features, while the decoder

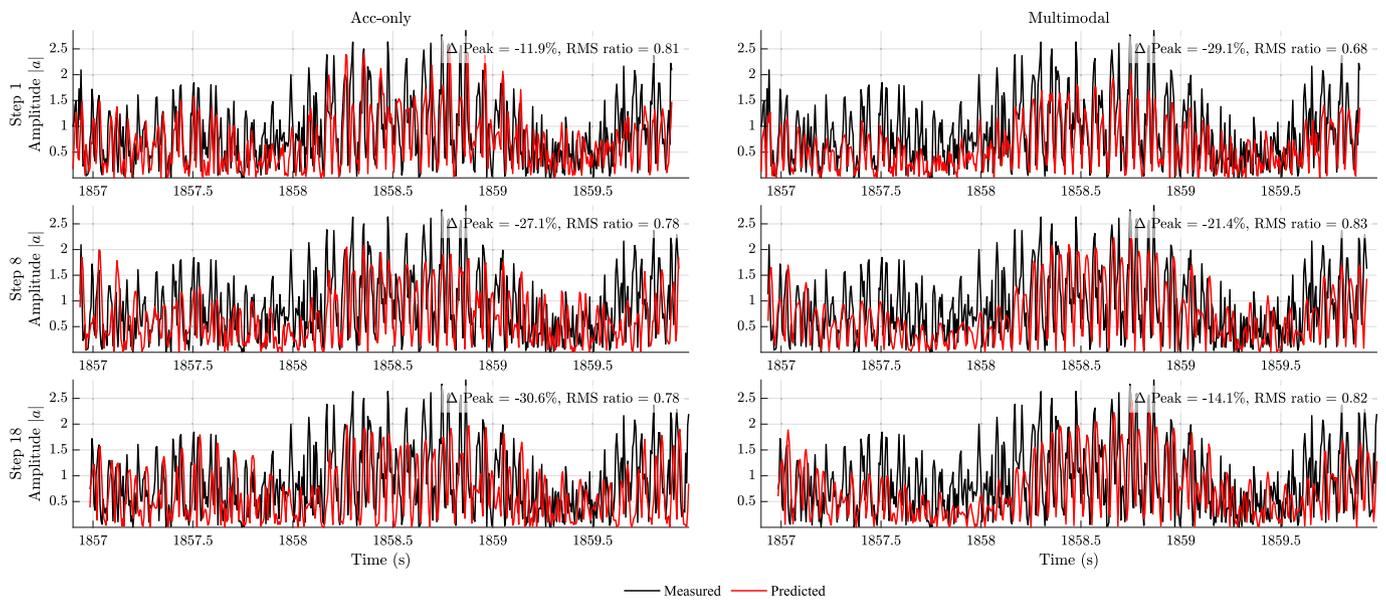

**Fig. 8.** Time-domain forecasting for $x$-axis (axial).

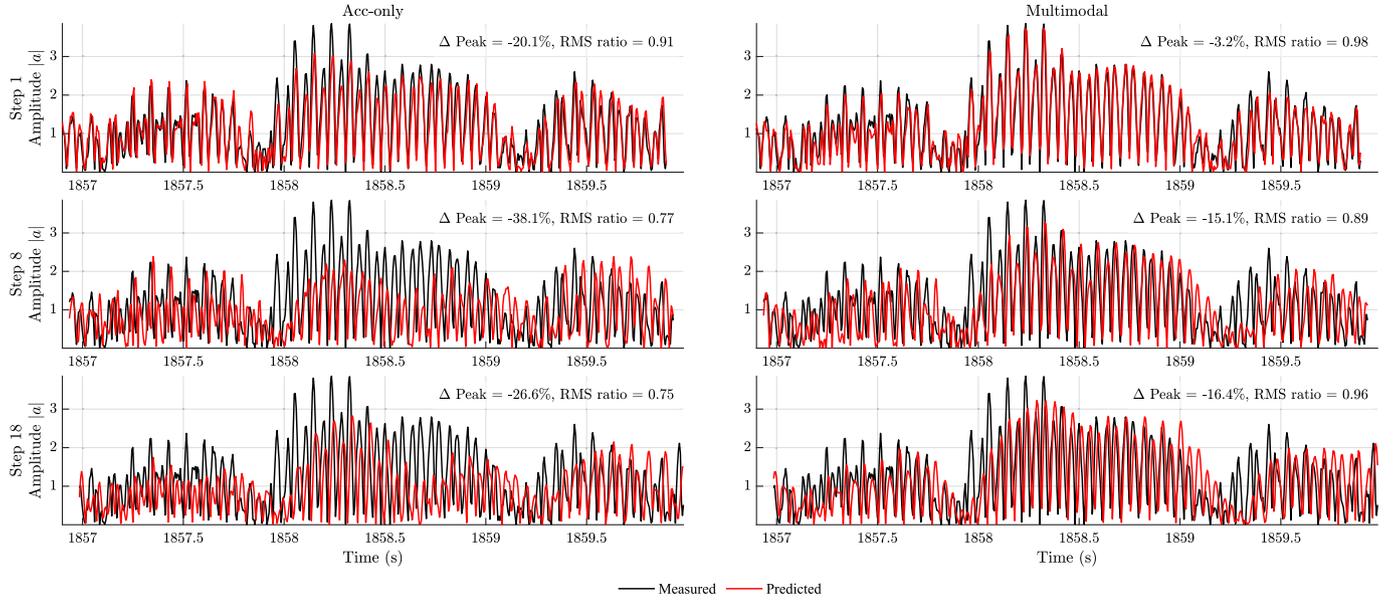

**Fig. 9.** Time-domain forecasting for *y*-axis (lateral).

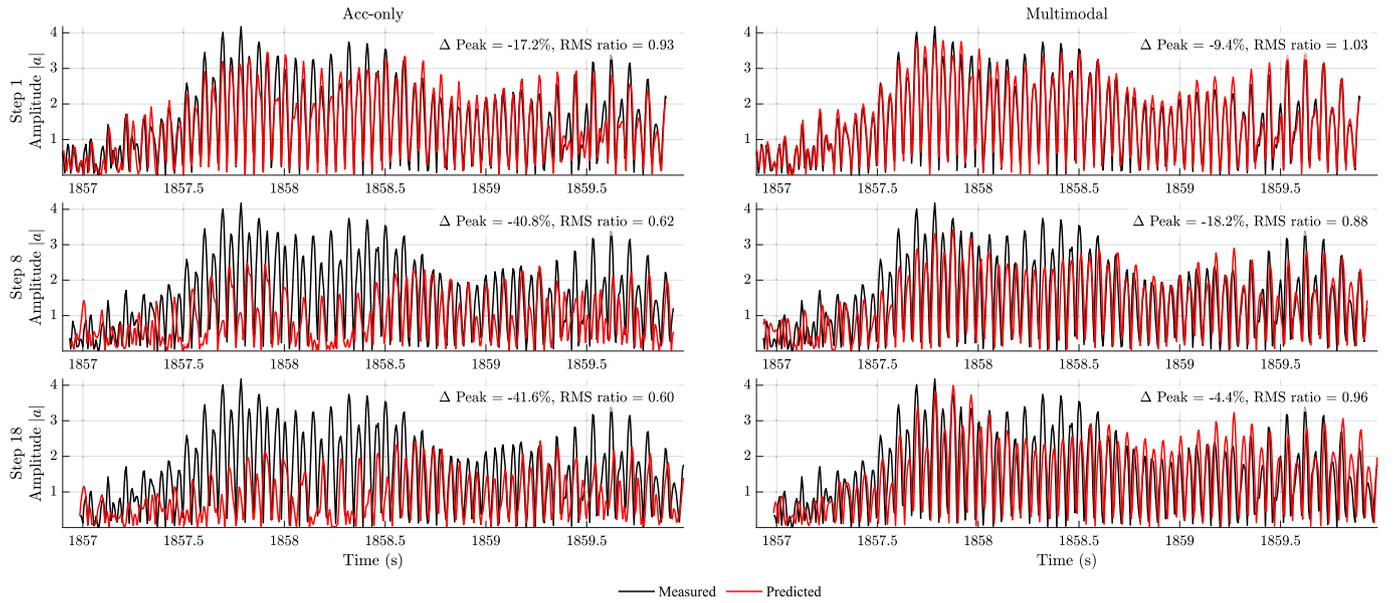

**Fig. 10.** Time-domain forecasting for *z*-axis (vertical).

cross-attention fuses wind and acceleration information by attending to the encoder memory $M$:

$$Z_{\text{dec}} = \text{softmax}\left(\frac{Q_{\text{dec}} K_{\text{enc}}^\top}{\sqrt{d_k}}\right) V_{\text{enc}} \tag{5}$$

The overall decoder output sequence $\hat{Y}$ is generated autoregressively by stacking $N$ layers of such attention and feed-forward submodules. Model parameters $\theta$ are optimized by minimizing the mean-squared error loss between predicted and measured accelerations:

$$\mathcal{L}(\theta) = \frac{1}{N L_{\text{pred}} d_a} \sum_{n=1}^{N} \sum_{t=1}^{L_{\text{pred}}} \|\hat{a}_{n,t} - a_{n,t}\|_2^2 \tag{6}$$

This formulation explicitly links the transformer's learned attention weights to the physical correspondence between wind excitation and structural response, enabling interpretable cross-modal forecasting in the early-warning setting.

## 4. Performance metrics for model validation

The model's forecasting performance for the structural response of all forecasting step scenarios is provided in this section. Mathematically, the $H$ forecasting-step at time $t$ is reported per axis $u \in \{x, y, z\}$ and per horizon $H \in \{1, 8, 18\}$. Namely, the model is examined for forecasting 1, 8, or 18 future steps ahead.

In the time domain, the peak error ΔPeak and the root-mean-square ratio (RMSR) are written as:

$$\Delta\text{Peak}^{(u)}(t; H) = 100 \cdot \frac{\max_{1 \leq h \leq H} \hat{y}_{t+h}^{(u)} - \max_{1 \leq h \leq H} y_{t+h}^{(u)}}{\max_{1 \leq h \leq H} y_{t+h}^{(u)}} \tag{7}$$

**Table 4**
Error comparison for acceleration-only and multimodal models.

(a) X-axis (longitudinal)

| Step | ΔPeak (%) A | ΔPeak (%) M | RMSR A | RMSR M |
|---|---|---|---|---|
| 1 | −11.9 | −29.1 | 0.81 | 0.68 |
| 8 | −27.1 | −21.4 | 0.78 | 0.83 |
| 18 | −30.6 | −14.1 | 0.78 | 0.82 |

(b) Y-axis (transverse)

| Step | ΔPeak (%) A | ΔPeak (%) M | RMSR A | RMSR M |
|---|---|---|---|---|
| 1 | −20.1 | −3.2 | 0.91 | 0.98 |
| 8 | −38.1 | −15.1 | 0.77 | 0.89 |
| 18 | −26.6 | −16.4 | 0.75 | 0.96 |

(c) Z-axis (vertical)

| Step | ΔPeak (%) A | ΔPeak (%) M | RMSR A | RMSR M |
|---|---|---|---|---|
| 1 | −17.2 | −9.4 | 0.93 | 1.03 |
| 8 | −40.8 | −18.2 | 0.62 | 0.88 |
| 18 | −41.6 | −4.4 | 0.60 | 0.96 |

$$\mathrm{RMSR}^{(u)}(t; H) = \frac{\sqrt{\frac{1}{H} \sum_{h=1}^{H} \left(\hat{y}_{t+h}^{(u)}\right)^2}}{\sqrt{\frac{1}{H} \sum_{h=1}^{H} \left(y_{t+h}^{(u)}\right)^2}} \tag{8}$$

where, max refers to the maximum vibration response of the structure for all time steps. Here, RMSR=1 denotes perfect power retention.

In the frequency domain, fixed modal bands are used for comparison of the ground truth and the model prediction. Let $B_k = [f_{L,k}, f_{U,k}]$ denote a band with fixed center $f_{0,k}$, and let $S_{yy}^{\mathrm{pred}}(f)$ and $S_{yy}^{\mathrm{meas}}(f)$ be the Welch power spectral densities of the prediction and measurement, respectively. Then, the band energy retention BER is provided as:

$$\mathrm{BER}_k = \frac{\sum_{f \in B_k} S_{yy}^{\mathrm{pred}}(f)\, \Delta f}{\sum_{f \in B_k} S_{yy}^{\mathrm{meas}}(f)\, \Delta f} \tag{9}$$

which equals one under perfect in-band power placement. Similarly, the modal peak error MPE at the fixed center is written as:

$$\mathrm{MPE}_k(\%) = 100\, \frac{S_{yy}^{\mathrm{pred}}(f_{0,k}) - S_{yy}^{\mathrm{meas}}(f_{0,k})}{S_{yy}^{\mathrm{meas}}(f_{0,k})} \tag{10}$$

A final metric is related to risk. The error residuals are written as $e_{m,t+h}^{(u)} = \hat{y}_{m,t+h}^{(u)} - y_{t+h}^{(u)}$ for either the acceleration-only scenario denoted by the subscript Acc, or the multimodal model scenario denoted by the subscript MM. The following quantities are then defined:

$$\Delta \mu = \mu_{\mathrm{MM}} - \mu_{\mathrm{Acc}}, \quad r_\sigma = \frac{\sigma_{\mathrm{MM}}}{\sigma_{\mathrm{Acc}}}, \quad \Delta p_{3\sigma_{\mathrm{Acc}}} = p_{\mathrm{MM}} - p_{\mathrm{Acc}}, \tag{11}$$

where $\mu$ and $\sigma$ are the mean and the standard deviation of $e$, while

$$p_m = \Pr\left(|e| > 3\sigma_{\mathrm{Acc}}\right) \tag{12}$$

Here, $\Delta p_{3\sigma_{\mathrm{Acc}}} < 0$ indicates the tail-risk reduction related to the fixed acceleration-only scenario.

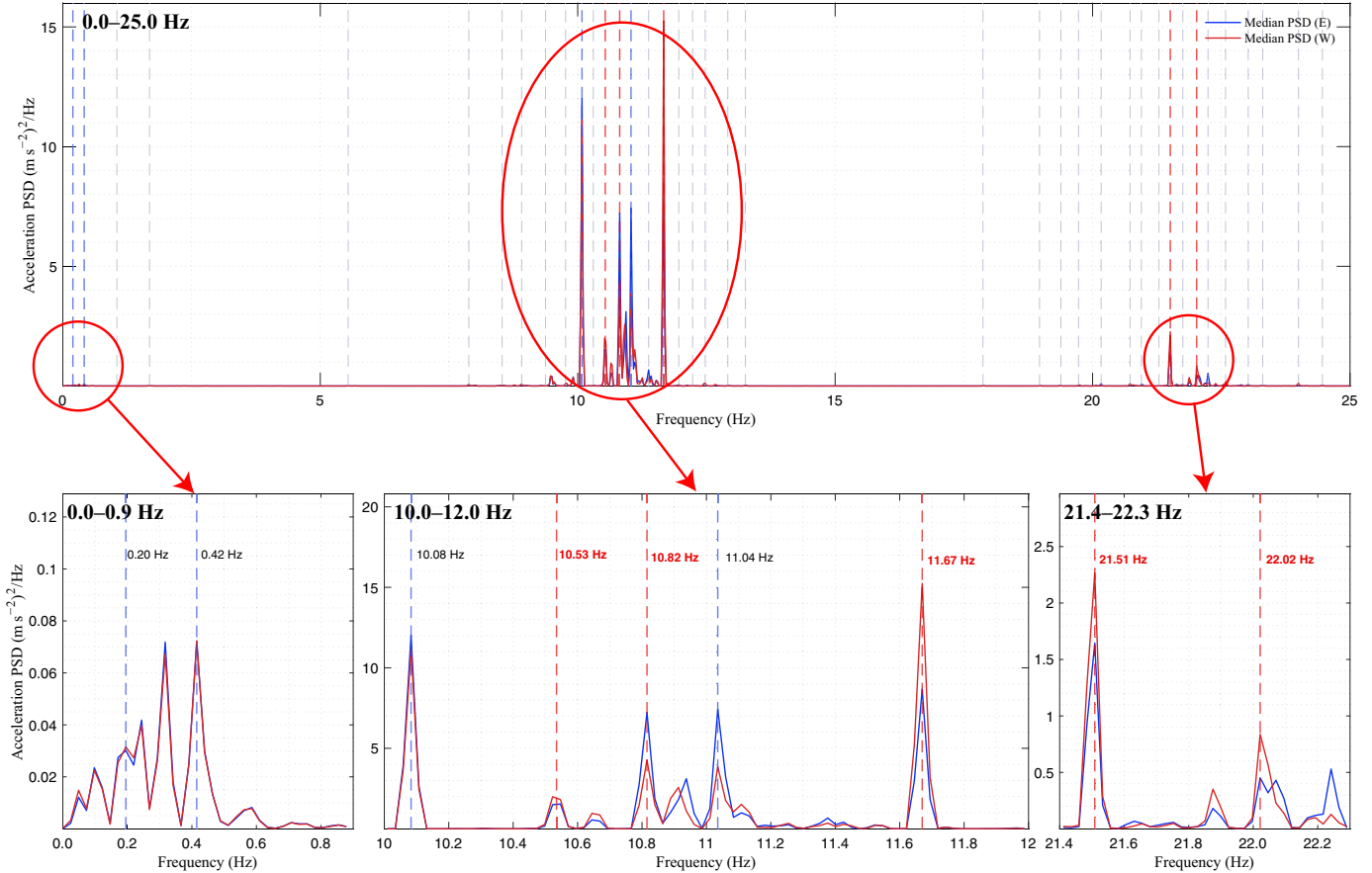

**Fig. 11.** Results and comparison for the median Welch power spectral densities on the East/West sides for the *y*-axis.

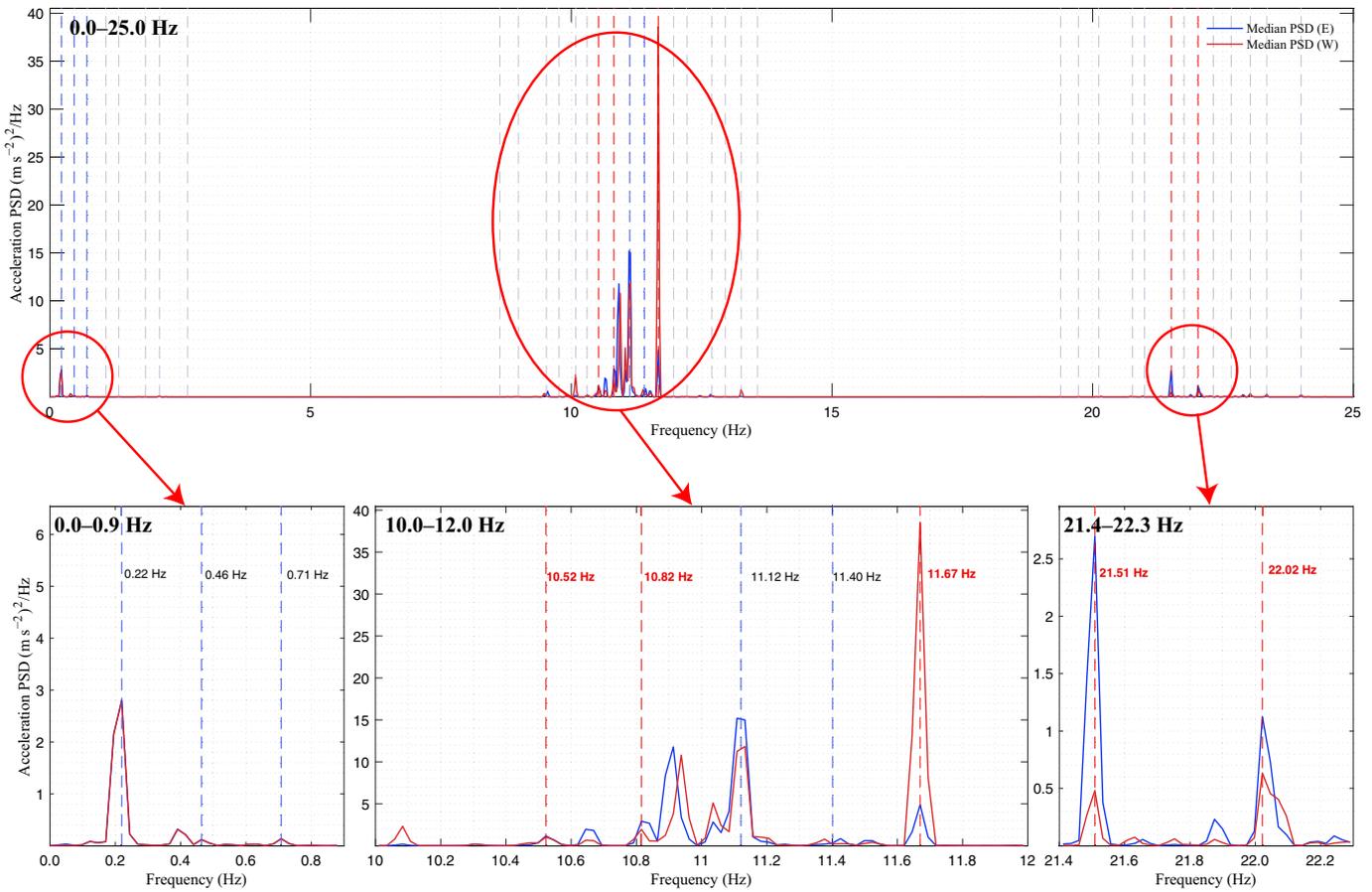

**Fig. 12.** Results and comparison for the median Welch power spectral densities on the East/West sides for the *z*-axis.

Finally, for any paired sequences $\{y_i, \hat{y}_i\}_{i=1}^{N}$, the root-mean-square error and mean absolute error are written as:

$$\text{RMSE} = \sqrt{\frac{1}{N}\sum_{i=1}^{N}(\hat{y}_i - y_i)^2}, \quad \text{MAE} = \frac{1}{N}\sum_{i=1}^{N}|\hat{y}_i - y_i| \quad (13)$$

while, the win rate for any chosen scalar $M$ such as RMSE or MAE is written as:

$$\text{Win\%}(M) = 100 \cdot \frac{1}{|A|} \sum_{a \in A} \mathbf{1}\{M_{\text{MM}}(a) < M_{\text{Acc}}(a)\} \quad (14)$$

with $A$ denoting the whole set. Importantly, all discussed metrics are computed per axis and step-horizon, and on identical windows.

## 5. Application to the hardanger bridge structural health monitoring system

The methodology is applied to the Hardanger Bridge structural health monitoring system, which contains long-term records of wind and deck acceleration [90]. The Hardanger Bridge is a 1308 m main-span suspension bridge in western Norway. Installed by the Norwegian University of Science and Technology shortly after the bridge opened in 2013, the structural health monitoring system is designed to monitor wind loading and structural response under the complex fjord terrain in Norway. The dataset from the structural health monitoring system provides synchronized measurements of ambient wind and the bridge vibrational response over many years of operation. The bridge's monitoring network includes a dense array of sensors mounted on the towers and the deck. In total, twenty tri-axial acceleration sensors (Canterbury Seismic CUSP-3D MEMS accelerometers) are deployed.

These accelerometers have a measurement range of ±4g, a low noise floor (130 dB SNR), and sample at up to 200 Hz. Two accelerometers are mounted at each tower top (on the east and west faces as seen in Fig. 6), and additional accelerometers are installed symmetrically on both sides of the steel girder along the span to capture vertical, horizontal, and torsional motions. The wind-monitoring subsystem consists of nine three-dimensional ultrasonic anemometers (Gill WindMaster Pro). Eight of these units are affixed to the bridge hangers along the main span, approximately 8 m above the deck to minimize flow disturbance, and one is placed on the top of the north (Vallavik) tower. Each Gill instrument measures 3D wind velocity (speed up to 65 m/s) and direction at up to 32 Hz. Finally, all sensor data are fed into local data-logger units via Cat-5 Ethernet. Each logger is equipped with a GPS receiver for precise time stamping and a Wi-Fi radio for communications. In this way, all sensor streams share a common clock and can be aligned on a global time base. An overview of the system (sensor locations, dataloggers, GPS units, etc.) is provided by Fenerci et al. [90].

In this work, measurements from the anemometers A2 and A4, and the deck-mounted accelerometers H3E and H3W of Fig. 6 are chosen for the adjusted "HighFreq" data within the database [90]. This case provides a local inflow characterization along the mid span via A2/A4, and collocated torsion-sensitive structural response through the H3E/H3W pair mounted on opposite sides of the steel box girder. The chosen subset, therefore, captures both the aerodynamic excitation and the coupled vertical—lateral—torsional deck dynamics in the same vicinity, facilitating cross-modal learning and validation on high-fidelity time histories. The examined horizon steps reported in the results (Steps 1/8/18) correspond to forecast horizons $h \in \{1, 8, 18\}$. Before feeding the data into the model, a preprocessing pipeline is applied to improve data quality and consistency as seen in Discussion, Tables 8 and 9, consisting of three stages: noise reduction, outlier handling, and normalization. Each recorded time series (wind or acceleration) is processed through these stages to yield clean, normalized data ready for model input. Fig. 7 illustrates the forecasting workflow, which includes the input and preprocessing stages, the Transformer encoder–decoder stage, and the residual energy module for anomaly detection.

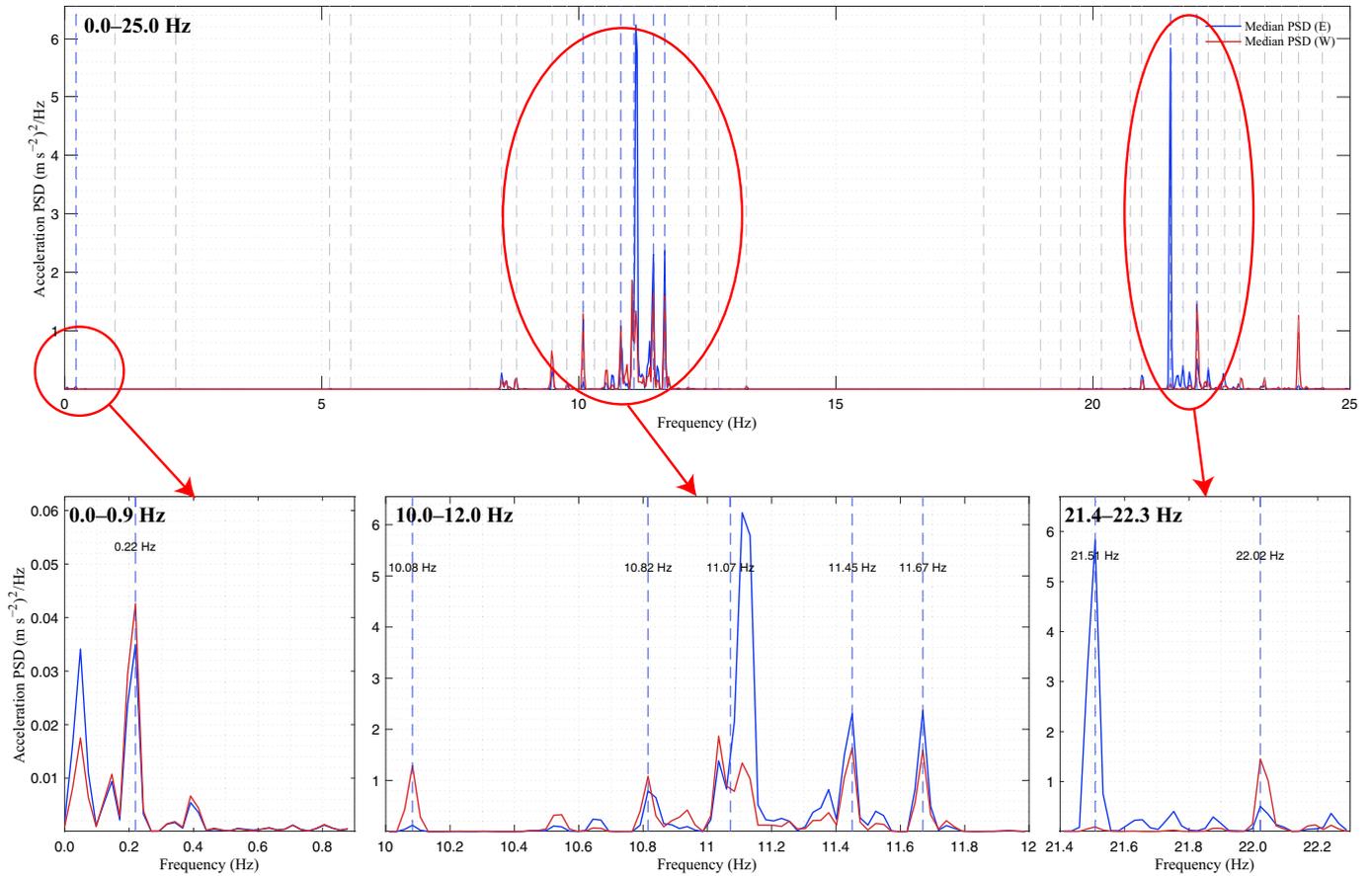

**Fig. 13.** Results and comparison for the median Welch power spectral densities on the East/West sides for the *x*-axis.

## 6. Examined transformer model results

To assess the performance of the model in time-domain, Figs. 8–10 are provided for both the acceleration-only and the multimodal forecaster, where the multimodal forecaster shows an improved performance. On the *x*-axis, the acceleration-only model underestimates the extreme vibration amplitudes at all step-horizons. On the *y*-axis, the multimodal model also shows a performance advantage at Steps 8/18. Finally, on the *z*-axis, this conclusion is even stronger, especially at medium/long step-horizons.

When it comes to the peak accuracy and normalized RMS error from Eqs. (7) and (8), Table 4(a)–(c) are provided. For the *x*-axis, the peak error of the multimodal model improves from −30.6% to −14.1% with RMSR 0.78 → 0.82. For the *y*-axis, the multimodal model also shows improved performance (e.g., Step 1: −3.2%, 0.98) and preserves higher energy at Steps 8/18. Finally, for *z*-axis, an even stronger conclusion is reached (Step 18 peak error −41.6% → −4.4%; RMSR 0.60 → 0.96).

In the frequency domain, the forecast and measured spectra are compared on the same Welch grid and within fixed modal bands, as defined in Eqs. (9) and (10). Intuitively, $BER_k \approx 1$ indicates perfect in-band energy retention (values <1 indicate attenuation; >1 over-energisation), while $MPE_k = 0$ indicates perfect accuracy at the fixed modal center $f_{0,k}$. All power spectral densities use identical Welch settings on a shared frequency grid, and fixed modal bands $B_k$ with centres $f_{0,k}$. Figs. 11–13 present these bands for the *y*, *z*, and *x* axes. Zoomed results are also provided in Figs. 14–16. Table 5 also summarizes the errors in the frequency domain.

Finally, the error residual distribution is shown in Fig. 17 and Table 7. Each subtable entry reports a bias shift $\Delta \mu = \mu_{MM} - \mu_{Acc}$, a variance ratio $r_\sigma = \sigma_{MM}/\sigma_{Acc}$, and a tail–difference $\Delta p_{3\sigma_{Acc}} = p_{MM} - p_{Acc}$ in percentage points, where $p_m = \Pr(|r_m| > 3\sigma_{Acc})$ defines the reference threshold at the acceleration-only $3\sigma$ per step. Along with the metrics in Table 6, it is shown that a greater concentration exists around zero-error, with an improved performance for the multimodal model, especially at the *y*/*z*-axes.

## 7. Discussion

This work provided a forecasting approach for the behavior of a structural system under wind excitation. The proposed methodology aimed to capture the temporal evolution of the system through a data-driven model that learns what is considered normal vibration behavior under varying wind and structural conditions. A major challenge in structural health monitoring under environmental variability is that the notion of "normality" is ill-defined when traditional methodologies struggle to distinguish between environmental or traffic, and true structural changes. In this study, the forecasting model identifies abnormal patterns by comparing the expected response (from the learned model) to the observed data. Deviations from the forecast can then be interpreted as potential signs of damage or change in the structural system, and with a wider application range from bridges, civil infrastructure, and beyond. The proposed framework is the first application of a multimodal encoder–decoder Transformer architecture for wind–structure interaction forecasting, jointly learning from synchronized wind and acceleration histories without requiring explicit aerodynamic or structural modeling. Unlike prior transformer-based structural monitoring studies that focus on vibration-based damage detection or single-modal vibration signals, the approach leverages cross-modal attention to capture causal dependencies between environmental excitation and structural response. The framework also operates under non-stationary conditions without assumptions of normal vibration behavior, addressing a major limitation of conventional physics-informed or statistical models.

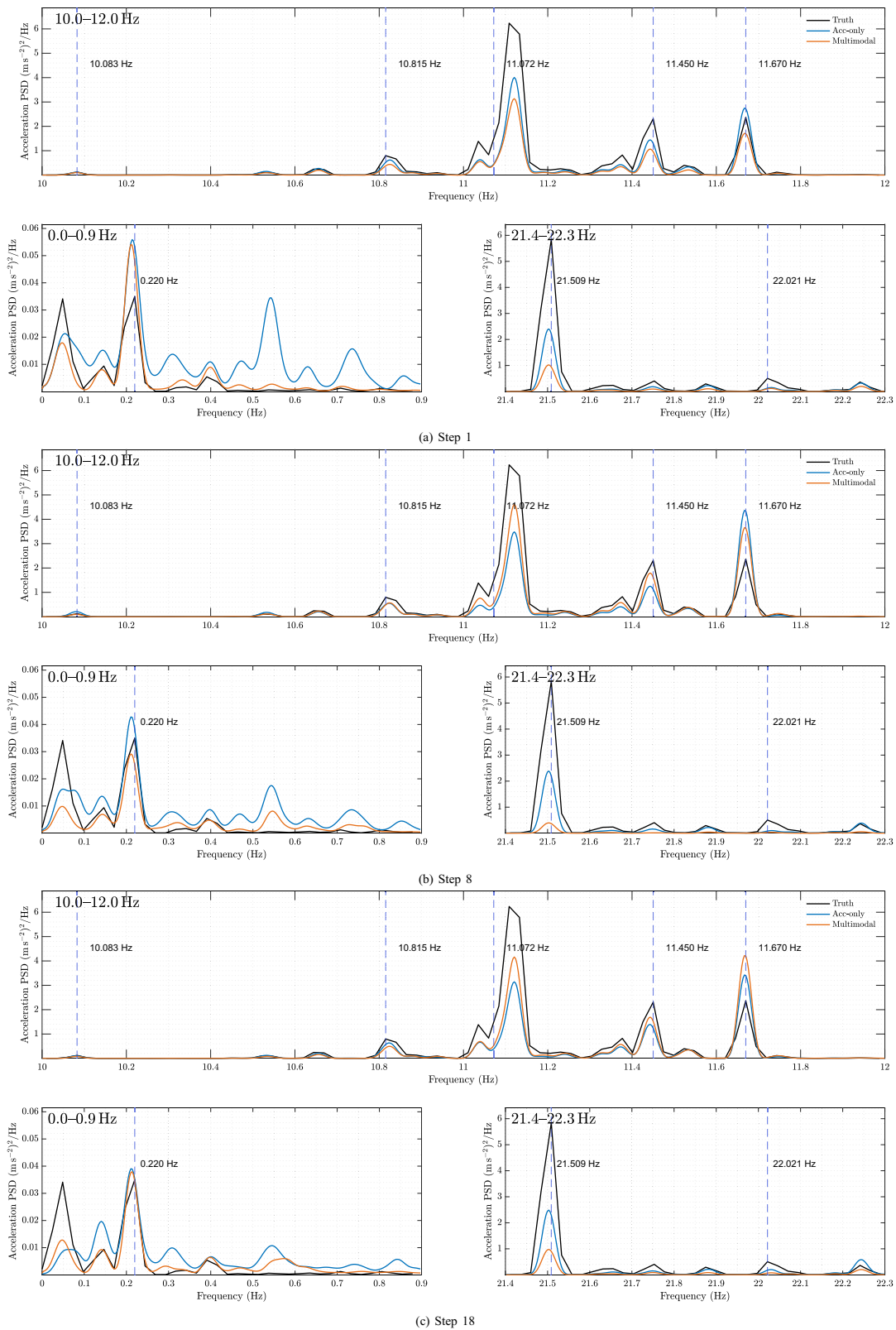

**Fig. 14.** Zoomed results and comparison for the median Welch power spectral densities on the East/West sides for the *x*-axis.

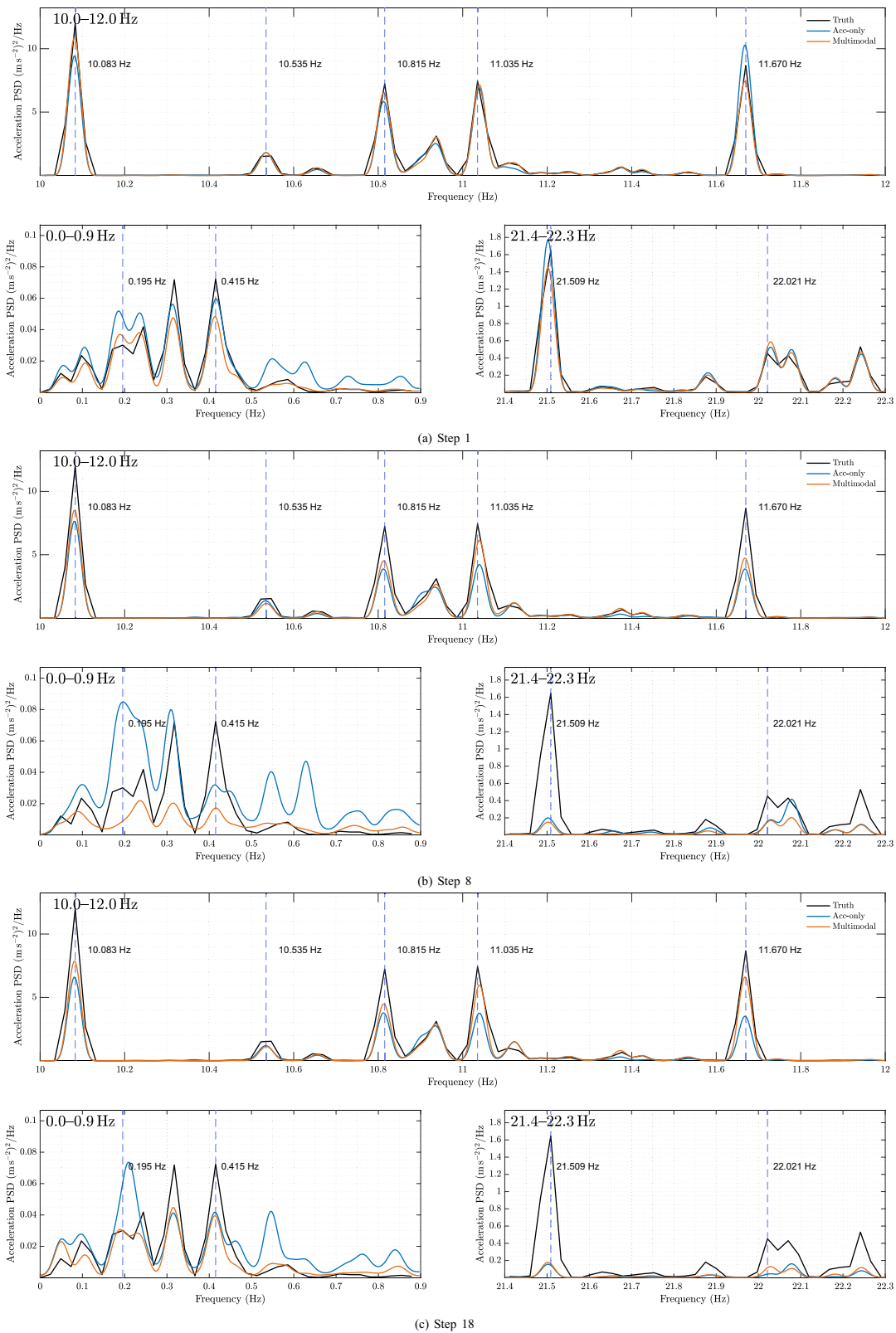

**Fig. 15.** Zoomed results and comparison for the median Welch power spectral densities on the East/West sides for the *y*-axis.

A concern is related to the performance of the model under different wind amplitude scenarios. To address this issue, Figs. 18 and 19 are also provided. They illustrate both the vibration response and the corresponding forecasting performance for the multimodal and acceleration-only models under low- and high-wind excitation conditions. The figures also depict the instantaneous prediction error at each time step for both scenarios. As observed, the low-wind case is associated with higher forecasting accuracy, as fewer nonlinear aerodynamic and structural effects dominate the response. Conversely, during strong wind excitation, the system exhibits more complex nonlinear interactions that are harder for the current model to capture accurately. Further research is therefore required to enhance model robustness under extreme wind events, which is essential for reliable early-warning structural health monitoring applications.

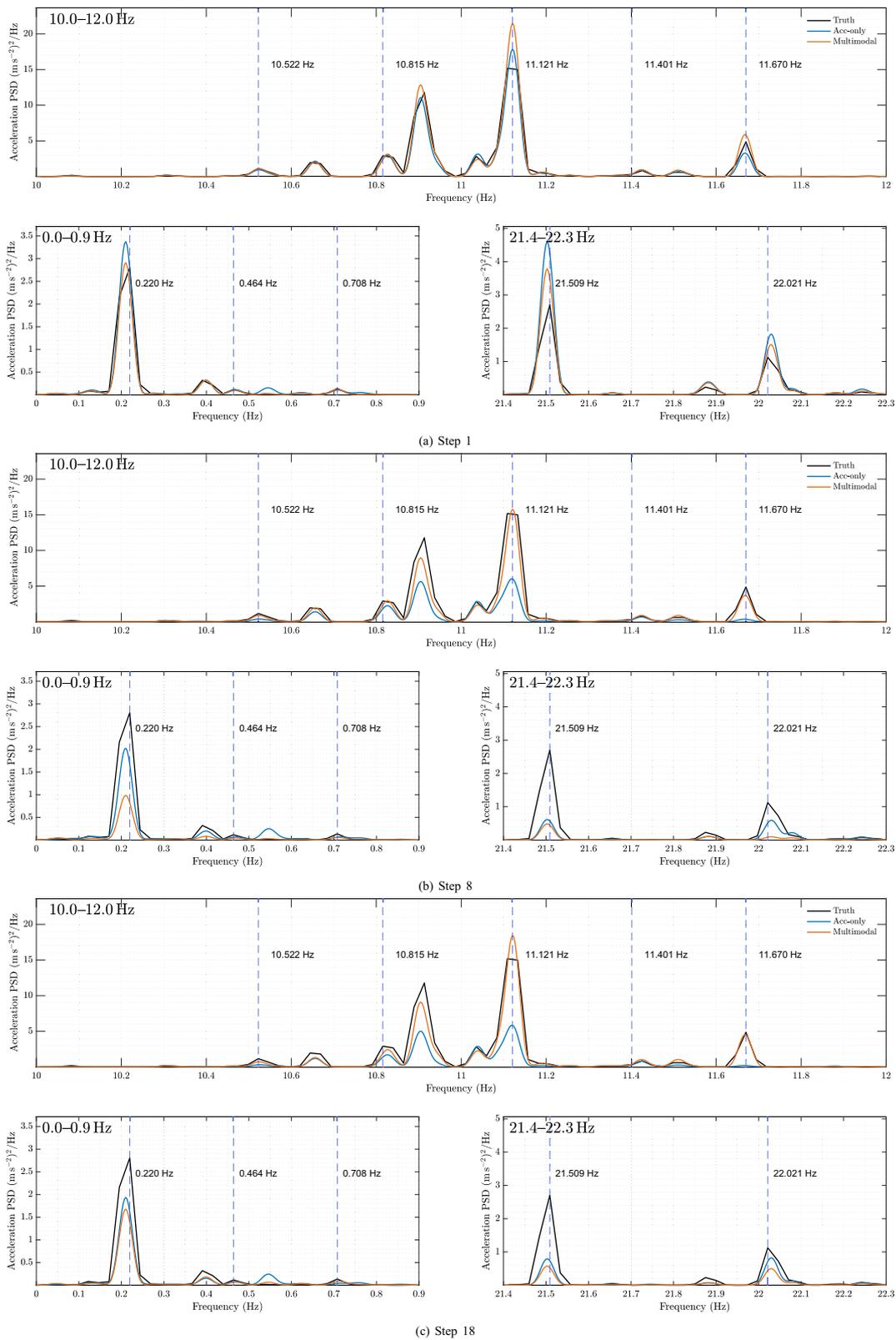

**Fig. 16.** Zoomed results and comparison for the median Welch power spectral densities on the East/West sides for the *z*-axis.

In this work, the architectural description of the proposed multimodal forecasting model is provided through Figs. 4 and 5. Since the proposed model is a Transformer–based encoder—decoder system, the architecture is naturally decomposed into two complementary components: Fig. 4 illustrates the encoder, which embeds and processes the wind modality to generate the memory tensor M, while Fig. 5 illustrates the decoder, which fuses past acceleration information with the encoder memory through masked self-attention and encoder—decoder cross–attention to generate multi-step acceleration forecasts.

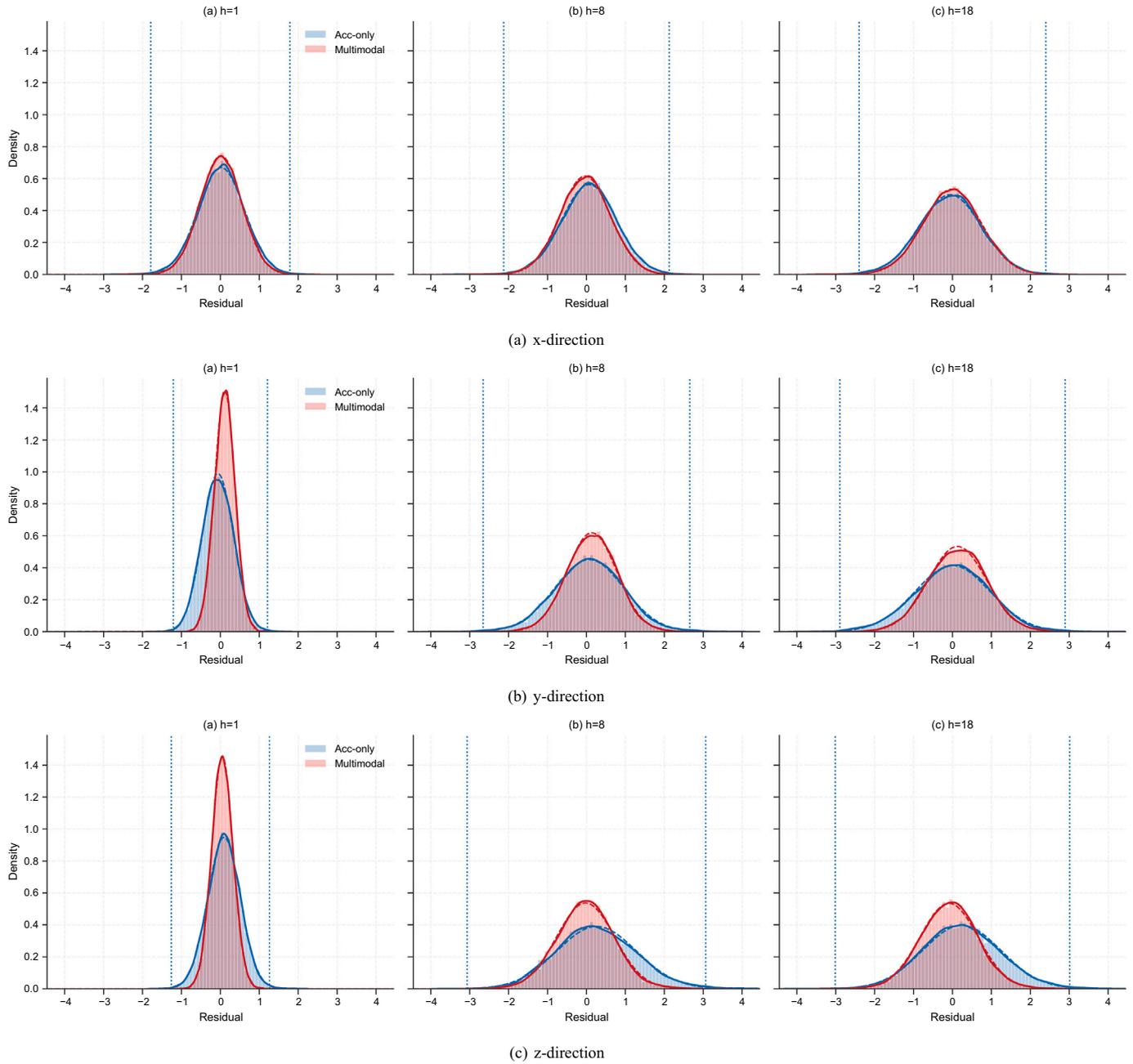

**Fig. 17.** Error residual–distribution for the last 20,000 samples across all steps and modes.

**Table 5**
Frequency-domain error comparison (median across shared $y/z$ core modes).

| Horizon | Model | Median $BER_k$ [–] | Median $MPE_k$ (%) |
|---|---|---|---|
| Step 1 | Acceleration-only | 0.944 | 17.1 |
|  | Multimodal | 0.897 | 14.8 |
| Step 8 | Acceleration-only | 0.503 | 56.8 |
|  | Multimodal | 0.553 | 45.1 |
| Step 18 | Acceleration-only | 0.467 | 58.8 |
|  | Multimodal | 0.578 | 42.2 |

To justify the Transformer modeling choice, a 1D convolutional neural network encoder—decoder baseline is also reported for the same forecasting task as seen in Tables 8 and 9. The convolutional neural network uses the same synchronized wind/acceleration input windows to predict future accelerations over the same horizon. A wind encoder (stacked temporal Conv1D blocks) compresses the wind history into a latent conditioning representation, while an acceleration decoder models the acceleration history with temporal convolutions and fuses the wind conditioning to produce the next-step predictions. Multi-step forecasts are obtained via an autoregressive rollout by feeding each predicted step

**Table 6**
Error comparison by $\mu$, $\sigma$, and $p(> 3\sigma_{Acc})$.

X-axis (longitudinal) error.

| | $\mu$ | | $\sigma$ | | $p(> 3\sigma_{Acc})$ [%] | |
|---|---|---|---|---|---|---|
| Step | A | M | A | M | A | M |
| 1 | 0.021 | 0.015 | 0.597 | 0.536 | 0.4 | 0.1 |
| 8 | 0.065 | −0.033 | 0.709 | 0.648 | 0.3 | 0.1 |
| 18 | −0.060 | −0.010 | 0.800 | 0.747 | 0.3 | 0.1 |

Y-axis (transverse) error.

| | $\mu$ | | $\sigma$ | | $p(> 3\sigma_{Acc})$ [%] | |
|---|---|---|---|---|---|---|
| Step | A | M | A | M | A | M |
| 1 | −0.063 | 0.111 | 0.403 | 0.264 | 0.3 | 0.0 |
| 8 | 0.074 | 0.138 | 0.885 | 0.645 | 0.5 | 0.0 |
| 18 | 0.001 | 0.127 | 0.965 | 0.748 | 0.3 | 0.0 |

Z-axis (vertical) error.

| | $\mu$ | | $\sigma$ | | $p(> 3\sigma_{Acc})$ [%] | |
|---|---|---|---|---|---|---|
| Step | A | M | A | M | A | M |
| 1 | 0.094 | 0.052 | 1.421 | 0.274 | 0.4 | 0.0 |
| 8 | 0.260 | −0.024 | 1.022 | 0.743 | 0.6 | 0.0 |
| 18 | 0.237 | −0.090 | 1.004 | 0.746 | 0.6 | 0.0 |

back into the decoder input, matching the iterative decoding protocol used in this work. All models are trained and evaluated under the same preprocessing, chronological split, normalization, and metric pipeline to ensure like-for-like comparison. The Transformer is shown to outperform more classical convolutional neural network approaches in all system dimensions x (Fig. 20), y (Fig. 21), and z (Fig. 22).

An important aspect of this methodology is that the model is trained using either acceleration data alone, or a combination of wind and acceleration measurements. It is essential to clarify that although acceleration is physically an output quantity of a dynamic system, it is used here as an input feature for the forecasting model. This choice enables the model to exploit temporal dependencies and correlations for predictive and anomaly detection purposes, rather than reproducing the system's physics directly.

An additional relevant concern is related to whether the model achieves its best possible performance given the available data. While incorporating additional sensors, such as strain gauges or displacement measurements, could theoretically enhance the model, current results indicate that the prediction error is already manageable. Both

**Table 7**
Error comparison by RMSE, MAE, and win rate (%).

X-axis (longitudinal) error.

| | RMSE | | MAE | | |
|---|---|---|---|---|---|
| Step | A | M | A | M | Win (%) |
| 1 | 0.597 | 0.537 | 0.472 | 0.428 | 55.6 |
| 8 | 0.712 | 0.649 | 0.565 | 0.517 | 55.2 |
| 18 | 0.802 | 0.747 | 0.638 | 0.595 | 55.1 |

Y-axis (transverse) error.

| | RMSE | | MAE | | |
|---|---|---|---|---|---|
| Step | A | M | A | M | Win (%) |
| 1 | 0.408 | 0.287 | 0.327 | 0.230 | 63.1 |
| 8 | 0.888 | 0.659 | 0.703 | 0.527 | 64.1 |
| 18 | 0.965 | 0.759 | 0.765 | 0.610 | 62.5 |

Z-axis (vertical) error.

| | RMSE | | MAE | | |
|---|---|---|---|---|---|
| Step | A | M | A | M | Win (%) |
| 1 | 0.431 | 0.279 | 0.342 | 0.221 | 68.2 |
| 8 | 1.055 | 0.743 | 0.833 | 0.584 | 65.0 |
| 18 | 1.032 | 0.751 | 0.815 | 0.591 | 63.4 |

the time-domain envelope and the dominant frequency content of the response are accurately captured. Therefore, although future research could explore multimodal sensing to improve robustness, no significant improvement is expected without substantial new data sources.

One of the key advantages of this approach is its focus on local effects. The model can be trained on measurements from a limited portion of the structure, without the need to develop or calibrate a full model of the entire bridge. This property is especially valuable for localized damage detection, where the objective is to identify and track small changes at specific locations rather than assessing the global dynamic behavior of the whole system. The validated results underscore the key advantage of the proposed multimodal Transformer framework: its ability to capture long-range temporal dependencies and handle non-stationary wind—structure interactions without relying on explicit physical modeling. Unlike traditional approaches that assume stationarity or require calibrated aerodynamic parameters, the Transformer learns cross-modal relationships directly from raw measurements. This capability is reflected in the quantitative improvements observed across all evaluation metrics. For instance, in the time domain, the multimodal model reduced peak error by up to 37 percentage points compared to the acceleration-only baseline (e.g., z-axis Step 18: −41.6% to −4.4%) and improved RMS ratio from 0.60 to 0.96, indicating near-perfect energy retention. Similarly, in the frequency domain, the model achieved a 28% reduction in modal peak error at longer horizons (Step 18: 58.8% to 42.2%) and preserved in-band energy with median BER values close to unity. Importantly, the proposed framework offers pathways for integration into operational structural health monitoring systems. Its lightweight, one-dimensional input design and efficient Transformer architecture make it suitable for real-time deployment on edge devices or centralized monitoring platforms. In practice, the model can continuously ingest streaming wind and acceleration data, providing short-horizon forecasts that serve as early-warning indicators for abnormal structural behavior. Furthermore, the attention-based mechanism enables adaptive retraining as new data become available, allowing the digital twin component to evolve alongside the physical structure throughout its lifecycle. This adaptability is critical for long-term structural monitoring, where environmental conditions, traffic loads, and structural properties change over time.

Another interesting observation is that the model captures both physical and non-physical frequency components of the measured signals. For instance, in the case of the bridge, frequencies above approximately 5 Hz are unlikely to be associated with meaningful structural dynamics; they may instead originate from sensor characteristics, mounting effects, or measurement noise. Nevertheless, the model reproduces these components faithfully, offering the flexibility to either filter them out for engineering interpretation, or retain them for a purely digital-twin representation of the sensing system itself. For damage detection purposes, the low-frequency range, around 1 Hz, remains the most relevant, as it reflects the fundamental dynamic modes of the structure.

A final concern is related to the integration of the proposed forecasting framework within a broader digital twin paradigm. The digital twin concept is a relatively new topic. It consists of four essential components: (a) the physical entity and the data collection from it, (b) the development of a virtual representation or simulation of this entity, (c) the continuous interaction and synchronization between the physical and virtual systems, and (d) the application and predictive capabilities enabled by this coupling. In this context, the present work aims to establish a forecasting model that aligns with the digital twin concept by learning and updating the system's dynamic behavior. Through continuous AI-driven model refinement, the virtual representation can evolve alongside the physical structure, enabling an increasingly accurate and adaptive future representation for both monitoring and predictive decision-making purposes. Despite its demonstrated strengths, the proposed framework has limitations. For instance, the forecasting horizon examined in this study is relatively short (up to 18 steps ahead), which restricts its applicability for long-term predictive decision-making. While the attention-based architecture is inherently adaptable, its generalization to other bridges or structural systems may require retraining or transfer learning, particularly when sensor layouts or dynamic characteristics differ significantly. Furthermore, although the model handles non-stationary inputs effectively, its performance under extreme wind events showed reduced accuracy compared to moderate conditions, indicating a need for further robustness enhancements. Finally, the approach relies on high-quality, synchronized multimodal data streams; in scenarios where such data are unavailable or incomplete, the benefits of cross-modal learning may diminish. Future work should explore domain adaptation techniques and extended forecasting horizons to improve scalability and resilience across diverse infrastructures.

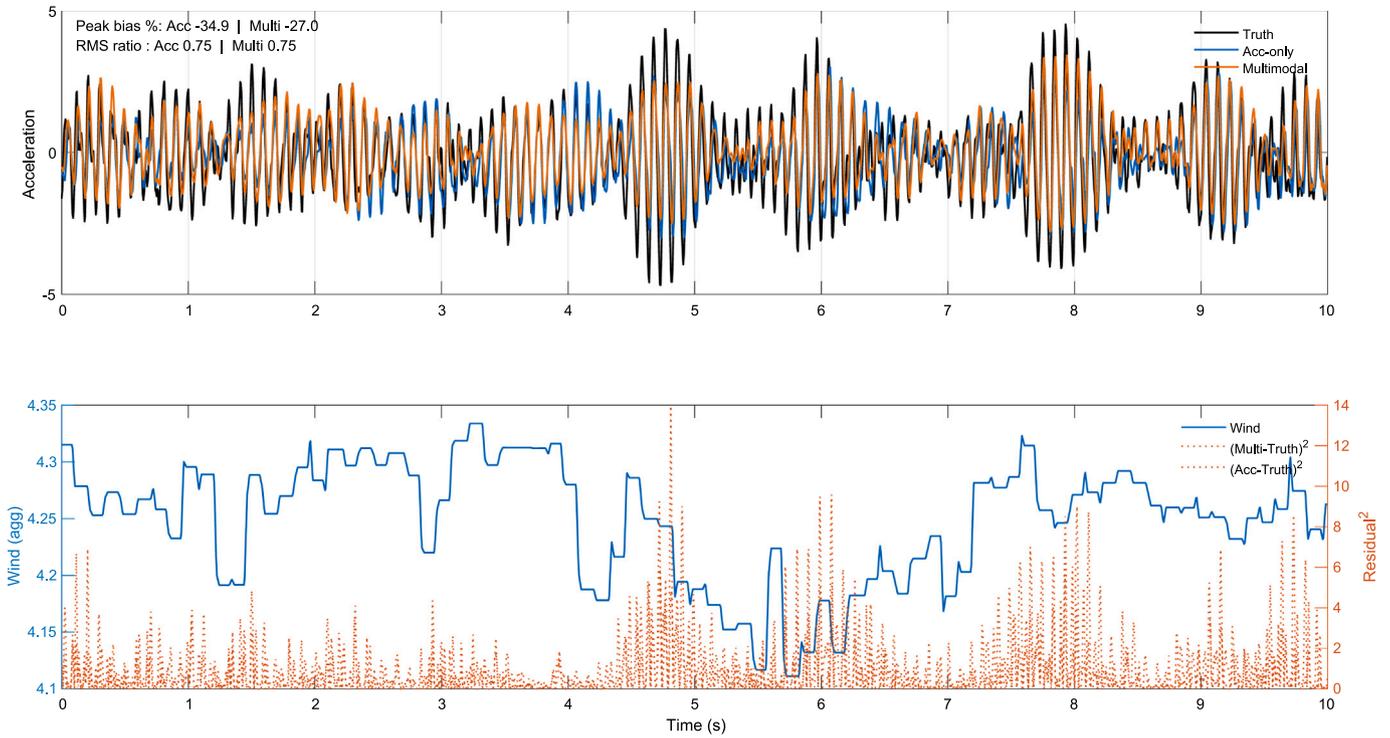

**Fig. 18.** Vibration response for the both the multimodal and the acceleration-only forecasting scenarios of Section 7 under strong wind excitation (upper plot), and corresponding instantaneous prediction error at each time-step (bottom plot).

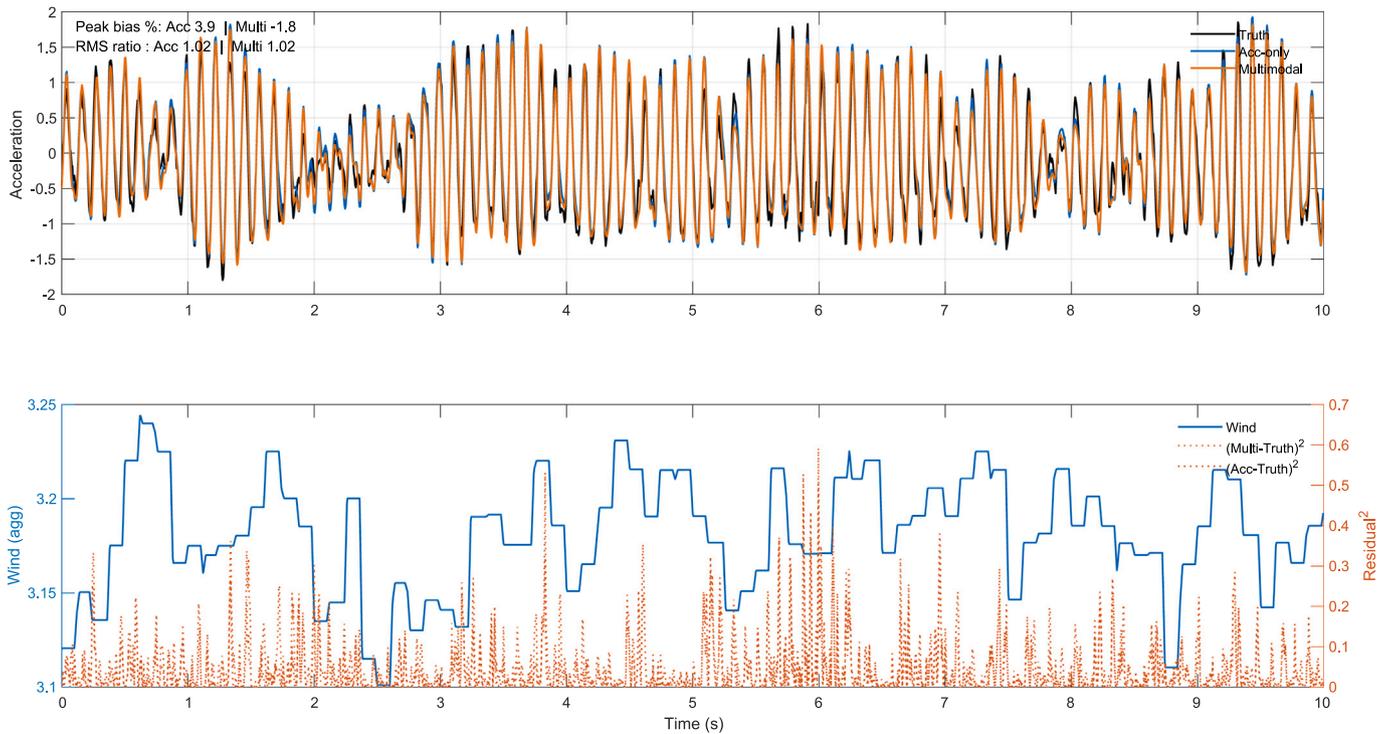

**Fig. 19.** Vibration response for both the multimodal and the acceleration-only forecasting scenarios of Section 7 under low wind excitation (upper plot), and corresponding instantaneous prediction error at each time-step (bottom plot).

**Table 8**
Model and training hyperparameters (RayTune-best reported runs).

| Hyperparameter | CNN (Wind + Acc) | Acc-only (Transformer) | Wind + Acc (Transformer) |
| --- | --- | --- | --- |
| Input dim (wind/acc) | 8/6 | 0/6 | 8/6 |
| Output dim (acc) | 6 | 6 | 6 |
| History length $L$/horizon $H$/stride | 100/20/1 | 100/20/1 | 100/20/1 |
| Model type | 1D CNN (Enc–Dec) | Decoder-only | Enc–Dec Transformer |
| Encoder layers/Decoder layers | 4/4 | 0/4 | 4/4 |
| Heads $n_h/d_{\text{model}}/d_{ff}$ | $d_{\text{model}} = 320$ | 8/256/384 | 8/256/384 |
| Positional encoding | none | sinusoidal | sinusoidal |
| Decoding | autoregressive | autoregressive + causal mask | autoregressive + causal mask |
| Loss | MSE | MSE | MSE |
| Batch size | 512 | 8 | 8 |
| Learning rate | $5.4010 \times 10^{-4}$ | $4.9720 \times 10^{-4}$ | $1.8663 \times 10^{-5}$ |
| Weight decay | $2.1948 \times 10^{-6}$ | $1.5191 \times 10^{-5}$ | $5.9663 \times 10^{-5}$ |
| Dropout | 0.0510 | 0.1315 | 0.2865 |
| Grad clip (norm) | 1.2071 | 0.5 | 0.5032 |
| Max epochs/early stop | 15/4 (best epoch = 9) | 100/5 | 100/5 |

**Table 9**
Validation protocol and evaluation settings (explicit, compact; applied consistently across all reported models).

| Item | Setting |
| --- | --- |
| Data streams | Synchronized wind and acceleration streams (two anemometers + east/west accelerometers), aligned by time stamps (row-wise alignment is used if an explicit `time` column is absent). |
| Split (chronological) | Temporally contiguous Train/Val/Test split (e.g., 70/15/15% by time); window start indices are constrained so that no input/target window crosses split boundaries (no leakage). |
| Windowing | Sliding windows with history length $L_{\text{enc}} = 100$ and prediction horizon $L_{\text{pred}} = 20$, stride = 1. Figures may display selected horizons (e.g., $H \in \{1, 8, 18\}$) while exports cover $H = 1 \dots 20$. |
| Normalization | Z-score normalization per feature using training-segment mean/std only (to avoid leakage); the same scaling is applied to validation/test. |
| Noise reduction (preprocessing) | Savitzky–Golay smoothing applied per channel (2nd-order polynomial, 7-sample window at $f_s = 200$ Hz), run forward–backward to avoid phase distortion. |
| Detrending/drift removal (evaluation) | Signals are high-pass filtered using a zero-phase 4th-order Butterworth filter implemented via forward–backward filtering (`filtfilt`), with cutoff $f_c = 0.05$ Hz. The same processing is applied to measured and predicted accelerations prior to metric computation (like-for-like). |
| Outlier/missing handling | Outliers are flagged if exceeding $5\sigma$ from a local running mean, or if the signal is constant (stuck/saturated) for $> 0.5$ s; flagged samples are repaired by linear interpolation between nearest valid points. Non-finite samples (NaN/Inf) are handled consistently in the same repair stage. |
| Time-domain metrics | RMSE and MAE vs horizon; peak bias $\Delta$Peak(%) and RMS ratio (RMSR) per axis and horizon on aligned (paired) windows. |
| Frequency/modal metrics | Welch PSD settings: Hann window 40 s, 75% overlap, $NFFT = 8192$ on filtered signals; fixed modal bands $\{B_k\}$ held constant across models/horizons. Peak picking uses minimum spacing 0.20 Hz and prominence relative to median noise floor; band edges via $-3$ dB half-power points; bandwidth clamped to $[0.05, 5.0]$ Hz. Metrics: BER, MPE, and OBER. |
| Residual/risk evaluation | Residual distribution analysis includes residual time series, histograms, QQ plots, and autocorrelation (ACF); tail summaries are reported via high quantiles of $|e|$ (e.g., 90/95/99%) on aligned windows. |
| Model selection | Transformer/Acc-only: best checkpoint selected by minimum *logged* validation loss in Ray Tune summaries. CNN baseline: hyperparameters selected by Ray Tune (ASHA) and then retrained for the final $L_{\text{pred}} = 20$ setting to match the Transformer horizon; the reported CNN checkpoint corresponds to the best validation-loss epoch. |

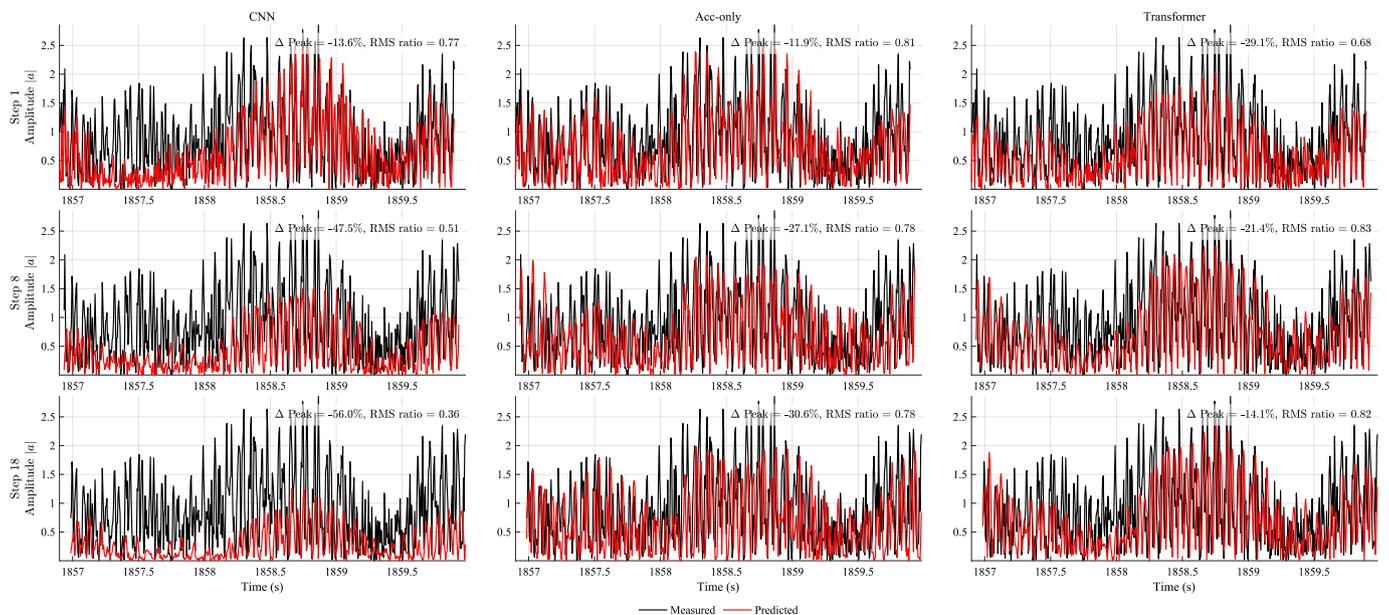

**Fig. 20.** Comparison of transformer and convolutional neural network approaches for *x*-axis (axial).

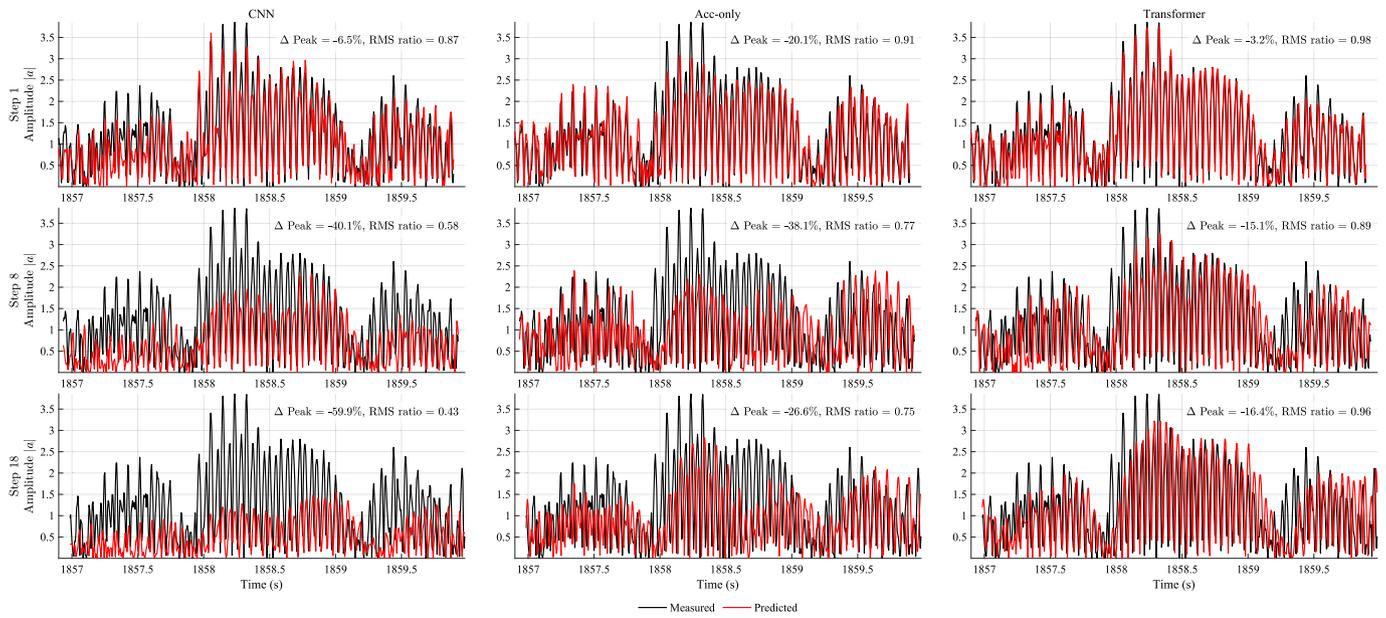

**Fig. 21.** Comparison of transformer and convolutional neural network approaches for *y*-axis (lateral).

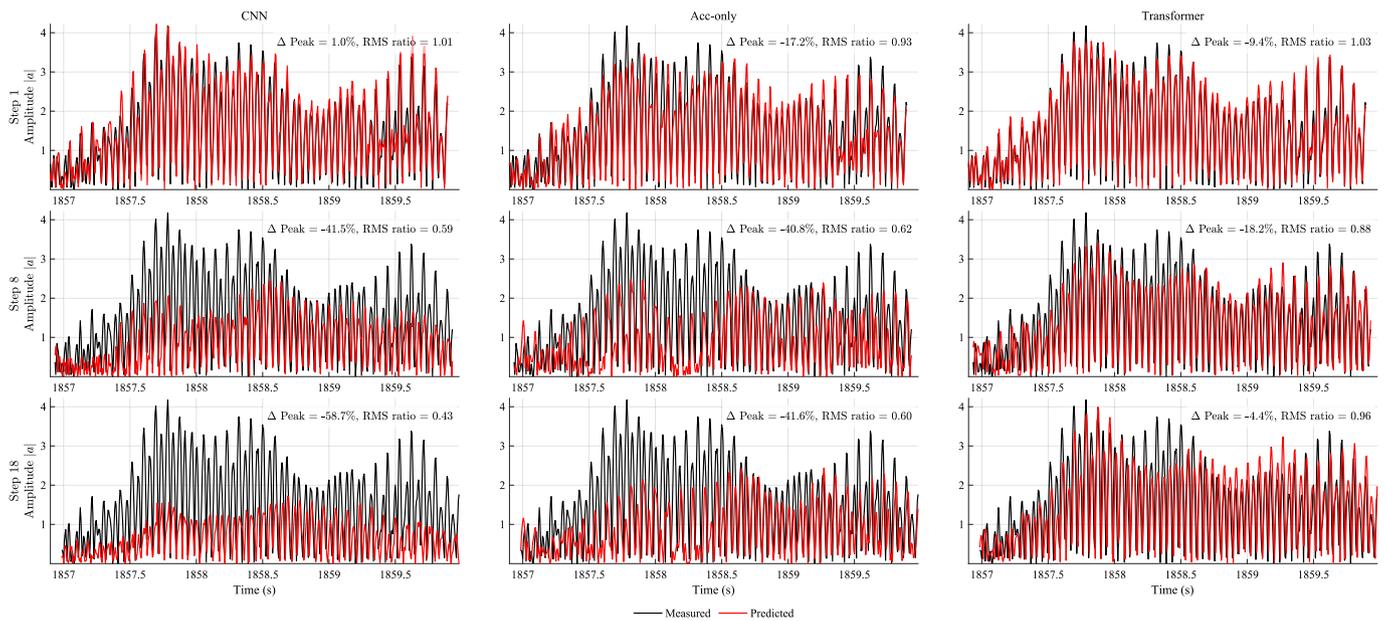

**Fig. 22.** Comparison of transformer and convolutional neural network approaches for *z*-axis (vertical).

## 8. Conclusions

The wind-excited structural response forecasting capabilities of a novel transformer model were examined. The framework was rigorously examined on real-world measurements from the Hardanger Bridge monitored by the Norwegian University of Science and Technology. The approach was shown to capture accurate structural dynamic behavior under wind excitation in realistic conditions, and with respect to the changes in the system excitation under different events. Specifically, the multimodal forecasting model, trained on both wind and acceleration measurements, reduced the amplitude attenuation observed in the acceleration-only scenario. It also captured the vibration peak values, and in the frequency domain, the multimodal model provided the correct in-band energy. Finally, the error residuals became more concentrated to zero values.

Overall, the approach allowed for structural response forecasting with:

1. No need to model the entire dynamic system.
2. Ability to capture long-dependent patterns.
3. Low-cost computation using one-dimensional measurements, suitable also for small-dataset scenarios.

4. A unique location application for localized damage or anomaly detection.
5. Independent of the system application.

Importantly, the examined approach provided a simple, yet effective and powerful, tool for monitoring systems and vibration-based damage detection applications in wind–structure dynamics with generality beyond bridges and civil infrastructure.

**CRediT authorship contribution statement**

**Feiyu Zhou:** Writing – original draft, Visualization, Validation, Software, Methodology, Investigation, Formal analysis, Data curation, Conceptualization. **Marios Impraimakis:** Writing – original draft, Writing – review & editing, Visualization, Validation, Supervision, Software, Methodology, Investigation, Formal analysis, Data curation, Conceptualization.

**Declaration of competing interest**

The authors declare that they have no known competing financial interests or personal relationships that could have appeared to influence the work reported in this paper.

**Acknowledgements**

The authors would like to gratefully acknowledge the Norwegian University of Science and Technology structural health monitoring group for providing the data, and the University of Bath for providing access to the computing resources of the dual-GPU workstation (2 × NVIDIA RTX 6000 Ada, 48 GB each, CUDA 12.4).

**Data availability**

All data used in this study are openly available at https://doi.org/10.21400/5ng8980s.